\documentclass[journal]{IEEEtran}
\usepackage[utf8]{inputenc}
\usepackage{color}
\usepackage[noadjust]{cite}
\usepackage[english]{babel}
\usepackage{amsmath}
\usepackage{amsfonts}
\usepackage{amssymb}
\usepackage{url}
\usepackage{graphicx}
\usepackage{multirow}
\usepackage{algorithm}
\usepackage{algorithmicx}
\usepackage{algpseudocode}
\usepackage{makecell}

\begin{document}

\title{One Shot Learning for Deformable Medical Image Registration and Periodic Motion Tracking}

\author{Tobias Fechter, Dimos Baltas%
\thanks{\textsuperscript{\textcopyright} 2020 IEEE.  Personal use of this material is permitted. Permission from IEEE must be obtained for all other uses, in any current or future media, including reprinting/republishing this material for advertising or promotional purposes, creating new collective works, for resale or redistribution to servers or lists, or reuse of any copyrighted component of this work in other works.}
\thanks{T. Fechter and D. Baltas are with the Division of Medical Physics, Department of Radiation Oncology, Medical Center – University of Freiburg, Faculty of Medicine. University of Freiburg,
Germany and also with the German Cancer Consortium (DKTK). Partner Site Freiburg, Germany (e-mail: tobias.fechter@uniklinik-freiburg.de, dimos.baltas@uniklinik-freiburg.de)}
\thanks{Corresponding author: Tobias Fechter}}

\maketitle

\begin{abstract}
Deformable image registration is a very important field of research in medical imaging. Recently multiple deep learning approaches were published in this area showing promising results. However,  drawbacks of deep learning methods are the need for a large amount of training datasets and their inability to register unseen images different from the training datasets. One shot learning comes without the need of large training datasets and has already been proven to be applicable to 3D data. In this work we present a one shot registration approach for periodic motion tracking in 3D and 4D datasets. When applied to a 3D dataset the algorithm calculates the inverse of the registration vector field simultaneously. For registration we employed a U-Net combined with a coarse to fine approach and a differential spatial transformer module. The algorithm was thoroughly tested with multiple 4D and 3D datasets publicly available. The results show that the presented approach is able to track periodic motion and to yield a competitive registration accuracy. Possible applications are the use as a stand-alone algorithm for 3D and 4D motion tracking or in the beginning of studies until enough datasets for a separate training phase are available.
\end{abstract}


\section{Introduction}

Deformable image registration (\textit{DIR}) is a technique to determine spatial non linear correspondences between two or more images and finds many important applications in the medical field. DIR can be used to detect anatomical changes within a patient (e.g. monitoring, breathing or heart motion) or to find correspondences among different patients (e.g. atlas based segmentation). Although, DIR has been an active research field for years, due to its complexity it remains a challenging problem. Some reasons that hamper the alignment of two or more images are: multiple image modalities, ill-posed high dimensional optimisation, different types of motion, bad image contrast \cite{sotiras2013}. These limitations require usually specialised algorithms tailored to a specific problem. Therefore, many innovative ideas have been proposed over the past few decades. Good overviews of the field can be found in \cite{sotiras2013,Keszei2016,Oh2017,Tavares2014, Tavares2014_2, Tavares2015}.

With the rise of machine learning in the recent years deep learning (\textit{DL}) based algorithms were presented to solve medical image registration problems. DL algorithms can learn relevant image features, the relations between features and how to translate the learned features into a deformation vector.

The best performing methods are based on convolutional neural networks (\textit{CNNs}). CNNs have an advantage in handling spatial information and are able to learn complex image features specific for the current registration problem.

One of the first approaches to align images with a CNN was the spatial transformer module \cite{NIPS2015_5854}, which was used to improve classification accuracy.

\cite{hu2018weakly} presented a CNN to register magnetic resonance images to intra-procedural transrectal ultrasound images of prostate cancer patients. They trained their network with labelled image datasets but after completion of training the network is able to register unlabelled images.

Other supervised methods were presented by \cite{YANG2017378} who showed a network for predicting the momentum of large deformation diffeomorphic metric mapping model, \cite{Fan2018} who presented a network for the registration of magnetic resonance images (\textit{MRI}) of the brain and \cite{Eppenhof2018, eppenhofSPIE} showed a method for pulmonary CT registration. 

The aforementioned methods can be classified as supervised methods. They need example registrations \cite{YANG2017378, Fan2018, Eppenhof2018} or manual segmentations \cite{hu2018weakly} for training the neural networks. Generating these information is complex, time consuming and problem specific. Recently, unsupervised methods \cite{Vos2017, Vos2019, Dalca2018, Krebs2019, Hongming2018} were presented. Similarly to conventional registration methods the authors of these methods find a solution to the registration problem at hand by optimising an image similarity metric.

One drawback of the above mentioned supervised and unsupervised DL methods is that a lot of training data are needed to optimise the network weights. Training data have to be selected carefully as the datasets should cover future registration tasks as best as possible. Therefore these algorithms cannot be applied to unseen domains \cite{Ferrante2018a}. Another machine learning approach that comes without the demand for big training datasets is one shot learning. Here the parameters of the network are trained from scratch only with the images to be registered. In \cite{Ferrante2018a} it has been shown that one shot learning can produce registration results comparable to state of the art methods for two dimensional image slices.

So far the presented \textit{CNN} based registration methods can only generate a deformation field for two three dimensional (\textit{3D}) image datasets. However, in some medical fields the application of four dimensional (\textit{4D}) data (3D + time) plays an important role. By way of example, organ or tumour movement due to respiratory motion is estimated by a 4D computed tomography (\textit{CT}) scan for radiotherapy treatment planning \cite{Maciejczyk2014} or the progress of a disease (e.g. tumour growth or shrinkage) during the course of treatment can be measured by follow up scans \cite{Oehlke2016}. To measure the growth of a tumour by means of a 4D dataset, DIR can be done by applying a 3D DIR algorithm to each 3D dataset and its temporal successor as the deformation of the tumour does not follow a specific pattern. Whereas, the motion of organs due to breathing or heartbeat is periodic and can not be covered by the multiple application of a 3D DIR algorithm.

For organ specific motion patterns \cite{Sundar2009} used a Gaussian kernel to calculate temporal smooth deformation fields for cardiac cine MRI datasets. B-splines with a periodic constraint were used by \cite{vandemeulebroucke2011spatiotemporal, Metz2011} for lung motion estimation. A trajectory constraint for periodic motion is presented in \cite{Peyrat2008, Peyrat2010}. The authors presented a multi-channel Diffeomorphic Demons based algorithm to register two 4D cardiac time series. In \cite{Wu2013} temporal coherence was ensured by smoothing temporal fibers. All these methods for the registration of 4D datasets rely on conventional image registration methods.

In this work we show a way how to use the strengths of CNNs to enable DIR of 4D periodic image datasets. We present an unsupervised CNN based DIR algorithm to calculate dense displacement fields for 4D image data with a periodic motion pattern. In case of 3D to 3D registration the algorithm calculates the deformation vector field (\textit{DVF}) and an approximation of the inverse field simultaneously. Registration is performed in one shot, so no training data are necessary. We evaluated the algorithm on multiple image datasets publicly available and are able to show that its performance is equal or superior to state of the art DL registration algorithms.


\section{Data}

The proposed algorithm was evaluated on 31 publicly available 4D datasets. Each of them having either manually set landmarks or manually drawn contours for registration evaluation. The 4D datasets in this work show organs with a periodic motion pattern and consist of multiple 3D datasets. Every 3D dataset (\textit{phase image}) shows the organ positions at a specific phase of the underlying motion. Registration results of the proposed algorithm were evaluated and compared to the results of other state of the art algorithms.

\subsection{DirLab}

To evaluate the presented algorithm in the thoracic region the publicly available DirLab \cite{Castillo2009, Castillo2009a} dataset was used. The dataset consists of 10 4D CTs of the thoracic region. Each 4D CT consists of 10 3D CTs and manually set reference landmarks in the lungs for registration evaluation. The full number of landmarks was identified only in the maximum inhalation and exhalation phase but in addition a subset of 75 features was marked on each of the expiratory phase images. In table \ref{tab:datasetDetails} more details can be seen.

\subsection{Popi}

Another dataset for evaluating image registration algorithms is provided by the Léon Bérard Cancer Center \& CREATIS lab, Lyon, France \cite{vandemeulebroucke2011spatiotemporal}. Six 4D CTs consisting of 10 3D CTs each, are provided. The datasets show the lung region and contain landmarks for registration evaluation. For the first three patients landmarks were identified in each of the 10 3D CTs, for the remaining three patients landmarks were identified in the CTs depicting the maximum inhale and exhale phase. Dataset details can be found in table \ref{tab:datasetDetails}.

\subsection{Sunnybrook}

To evaluate the performance of our algorithm on a different body site and modality we used the cardiac \textit{MRI} datasets \cite{Radau2009} provided by the Sunnybrook Health Sciences Centre. In particular, we used the 12 datasets that were also provided as training data for the MICCAI left ventricle segmentaion challenge \cite{MiccaiLVChallenge}. The datasets contain cine-MRI images from patients with different pathologies and expert-drawn contours for validation. Each datasets consists of 20 phase images and expert drawn contours are available for  end-diastolic and end-systolic phases. In table \ref{tab:datasetDetails} details on the datasets are listed (the datasets are abbreviated as \textit{SC-*}).

\section{Methods}

Let $I^N$ be a 4D medical image dataset consisting of $N$ 3D images (for conventional 3D to 3D DIR $N$ is 2). The aim of our deformable DIR algorithm is to find a transformation $T^N$ that maps each 3D image $I_n$ in $I^N$ to its timely adjacent successor $I_{n+1}$ (as we deal with periodic data $I_{N-1}$ is aligned to $I_0$). $T^N$ consist of $N$ 3D dense vector fields $u_n$ with 3 components (x-, y- and z-direction) each and describes the trajectory for each voxel in $I_n$. $T_i^j(x)$ is the composition and application of the deformation fields $T_i$ to $T_j$ to position $x$. The deformation vector at a given position $x$ and phase image $n$ is indicated as $u_n(x)$. 

In this work we assume that the images are pre-registered either by the tomography scanner or by using affine registration algorithms \cite{elastix2010, plastimatch}. The task at hand is to calculate the deformable transformation $T^N$ that represents anatomically plausible spatial relationships between voxels in $I^N$ and takes care of the periodic character of the underlying motion.

\subsection{Preprocessing}

To reduce the amount of data to be processed every dataset was divided into foreground and background region and only the foreground region was processed by the registration algorithm. For the Popi and DirLab datasets the foreground was defined as all voxels inside the body outline contour. The body outline contour was generated automatically with \cite{FECHTER2017S757}. For the Sunnybrook datasets the foreground was generated by discarding the largest connected region containing only voxels with intensity 0.

The Sunnybrook datasets have much smaller voxel spacing along the x- and y-axis than along the z-axis, which had a negative impact on the registration results. Therefore these datasets  were resampled to a voxel size of 1.5 mm x 1.5 mm x 1.5 mm. For the other two dataset collections this step was neglected because the differences in voxel spacing along the axes were not as big as for the Sunnybrook datasets and preliminary tests showed that resampling the datasets leads to no significant change in registration performance.

In a last preprocessing step the foreground voxel intensities were normalised to zero mean and standard deviation of one. 

\subsection{Network Architecture and Training}

\begin{figure*}[htb!]
     \begin{center}
     \includegraphics[width=0.85\linewidth]{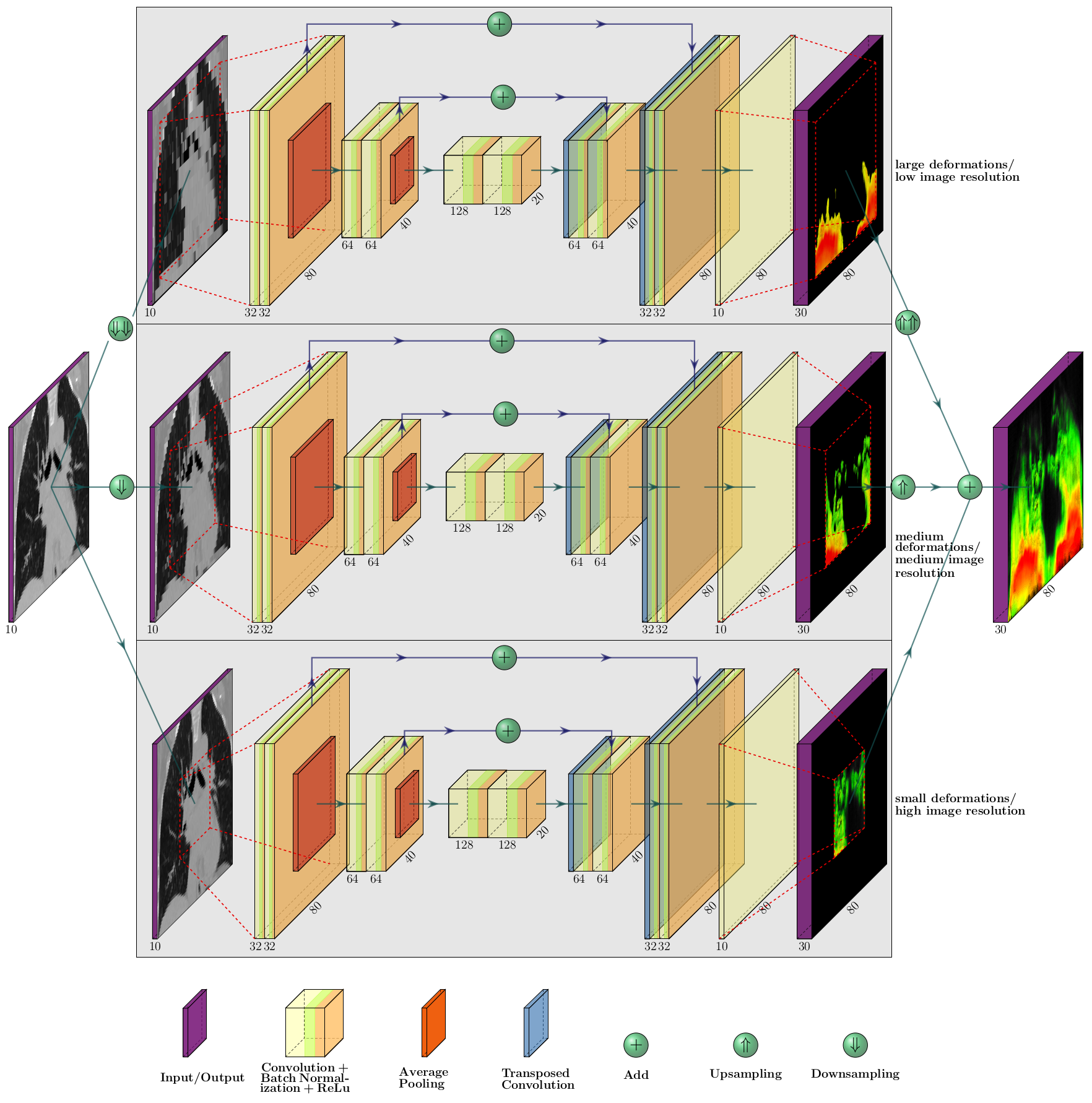}
     \caption{This figure shows the multi-resolution network architecture used in this work exemplarily for a 4D CT scan of the thoracic region with 10 phase images. The gray shaded rectangles stand for processing the input data in a different image resolution. To calculate large deformations (indicated in red in the output image) the U-net is trained till convergence on the two times downsampled input dataset. Then, medium and small deformations (yellow and green) are calculated by training the net on the simple downsampled and the original input dataset, respectively. The number on the lower left end of the squares indicates the number of channels, the number on the lower edge indicates the channel size. Note that the U-Net architecture is the same for all image resolutions, solely the input is changing. Convolutional layers had a kernel size of $3\times3\times3$, stride and padding were set to 1, the pooling layers had a kernel size of $2\times2\times2$ and a stride of 2. One downsampling step reduced the image size by a factor of 2.}
     \label{fig:architecture}
     \end{center}
     
\end{figure*}

Following the recently published works about DL based image registration \cite{Ferrante2018a, Vos2017, Vos2019, Dalca2018, Krebs2019, Hongming2018} we decided for a CNN based architecture. Similar to \cite{Ferrante2018a} we employed the U-net architecture \cite{Ronneberger2015} with 2 downsampling steps with average pooling layers, 2 upsampling steps with transposed convolution layers \cite{Dumoulin2016} and skip connections by summation instead of concatenation. The network  takes $I^N$ or a patch of $I^N$  as input (each phase image as a single channel) and calculates $N$ deformation fields $T^N$ with the network parameters $\Theta$. An illustration of all layers and connections can be found in Fig. \ref{fig:architecture}.

The idea of one shot learning in image registration is to start with an untrained model and to optimise the parameters until convergence only with the one dataset to be registered \cite{Ferrante2018a}.

\subsection{Multi-Resolution Patch Processing}

The large amount of memory which is needed for 4D image data, the corresponding deformation fields and the network parameters prevented to process datasets as a whole. Therefore, before processing an input dataset it is split into smaller non overlapping patches. Splitting is performed along the spatial dimensions only. The network is then trained till convergence with each patch. After termination of training the computed vector field patches are assembled.

The combination of patch based processing and the receptive field of $44\times44\times44$ voxels limits the ability of the presented network to account for large deformations (occurring e.g. in the diaphragmatic region). Thus we adopted a coarse-to-fine approach like it is often used in conventional image registration \cite{Schnabel2001}. Our multi-resolution approach consists of 3 image registration steps. First, basic deformations are calculated with the input image downsampled by a factor of 4. This step is repeated with the input image downsampled by a factor of 2 to capture intermediate deformations. In a last step the original image is fed to the network. The calculated vector fields of a respective downsampling step are added to the upsampled vector fields of previous steps before the loss $L$ is calculated. Thereby, the net first calculates a basic deformation field and adds more and more details after every upsampling step. The whole multi-resolution architecture is depicted in Fig. \ref{fig:architecture}.

\subsection{Loss Function}

\begin{figure}[htb]
     \begin{center}
     \includegraphics[width=0.8\linewidth]{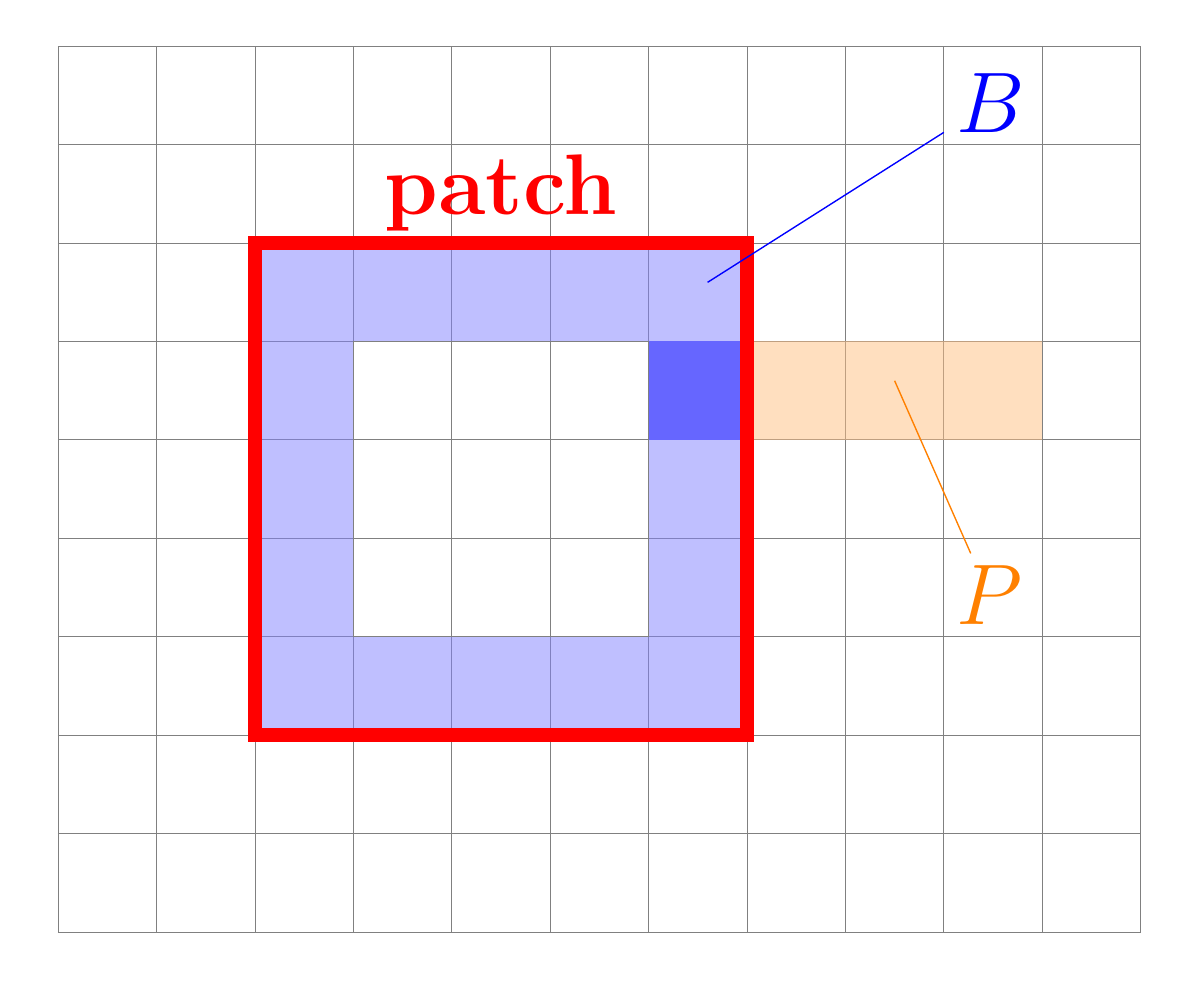}
                         
     \caption{The voxel sets needed for the calculation of the regularisation constraint $R$ are illustrated in this figure. The red line indicates a fictive image patch of 5 voxels. The area filled in blue represents the set of border voxels $B$ and the orange filled area $P$ contains the neighbouring voxels of  the dark blue coloured voxel along the x-axis.}
     \label{fig:smoothnessLoss}
     \end{center}
     
\end{figure}

To find $T^N$ we optimise $\Theta$ to minimise a loss function $L$. The value of $L$ indicates the quality of the deformation field and how well $T^N$ aligns the images in $I^N$. In this work $L$ is defined as:

\begin{equation}
\label{eqn:LossFunction}
L(I^N, T^N) = D(I^N, T^N(I^N)) + \lambda_0 R(T^N) + \lambda_1 C(T^N),
\end{equation}

and consists of 3 parts: an image dissimilarity metric $D$, a regularisation constraint $R$ for smoothness and to prevent folding as well as a cyclic constraint $C$ to encourage periodic deformations. $\lambda_0$ and $\lambda_1$ are weighting terms to control the influence of the regularisation terms. For $D$ the negative normalised cross correlation was adopted as it is known to perform well for mono modal image registration \cite{Roche2000}. 
$R$ was chosen to be the l2-norm derivatives of the deformation field in combination with a boundary smoothness constraint to guarantee a homogeneous deformation field also at patch transitions:

\begin{equation}
\label{eqn:smoothnessLossFunction}
\begin{split}
	R(T^N) = \sum_{x \in V}\sum_{n=0}^{N-1} \| \nabla u_n(x) \|_2^2 + \\ 
	\alpha \sum_{n=0}^{N-1}\sum_{x \in B}\sum_{y \in P} \dfrac{1}{d(x,y)} \| u_n(x) - u_n(y) \|_2^2,
\end{split}
\end{equation}

with $V$ being the set of all 3D voxel positions, $N$ the number of phase images, $B$ the set of voxel positions at the border of an image patch and $P$ the set of neighbouring voxel positions along a given axis. For a better understanding we visualised the sets $B$ and $P$ in Fig. \ref{fig:smoothnessLoss}. $\alpha$ controls the influence of the boundary smoothness constraint and $d(x,y)$ calculates the Euclidean distance between positions $x$ and $y$.

The periodic constraint $C$ was defined as the sum of all deformation vectors along the trajectory of a voxel:

\begin{equation}
\label{eqn:CycleLossFunction}
	C(T^N) = \sum_{x \in V}\sum_{n=0}^{N-1} u_n(T_0^n(x)).
\end{equation}

A sum of zero indicates a perfect periodic motion. To calculate $T^N(I^N)$ in (\ref{eqn:LossFunction}) and $T_0^n(x)$ in (\ref{eqn:CycleLossFunction}) we implemented a differential spatial transformer module \cite{NIPS2015_5854} that warps the input images and allows backpropagation of the loss to update $\Theta$. 

\subsection{Implementation}

The presented algorithm was implemented with pytorch v1.0.1. To calculate the gradients we made use of the pytorch autograd library which keeps track of all operations and builds a computational tree. However, for the implementation of the periodic constraint $C$ a workaround was necessary, as the library cannot backpropagate gradients through the compositon of DVF. We bypassed this problem by deforming an index tensor detached from the computational graph. As a result the gradients need to be calculated only for the sum of vectors referenced by the index tensor. For further details please see our code provided on GitHub \footnote{https://github.com/ToFec/OneShotImageRegistration}. Computations were done on a PC with a 8 core Intel Xeon CPU, 60 GB memory and a Nvidia Titan XP GPU. To simplify the adjustment of the loss function weights we normalised all parts of the loss function to a value range between 0 and 1. Like in the literature \cite{Vishnevskiy2014, Ferrante2018a, Vos2019} we set the image dissimilarity term to have the biggest impact and the regularisation parts only a fraction of it. In our test setting the parameters were set empirically to: $\lambda_0$ : $1\times10^{-3}$, $\lambda_1$ : $1\times10^{-2}$, $\alpha$: zero in the first multi-resolution step and 0.1 for the following steps. A sensitivity analysis that examined the robustness of our method against changes in the parameter set showed that varying $\lambda_0$ or $\lambda_1$ by a factor of 10 changes the average registration error by 0.13 mm. We assumed convergence when the current value of the loss function (\ref{eqn:LossFunction}) and the moving average of previous loss values did not differ by more than $1\times10^{-5}$.

\section{Evaluation}
\label{sec:evaluation}

For the datasets used in this study either manually set landmarks or contours were provided for registration quality assessment. To evalute the datasets with provided landmarks we calculated the distance between corresponding landmarks before and after registration. To evaluate datasets with manually drawn contours we made use of 3 methods usually used in the field of image segmentation. First, we investigated volume similarities by the S\o{}rensen-Dice index (DSC) \cite{sorensen1948method}.
Volume-based metrics alone might miss clinically relevant differences between contours as they show a lower sensitivity to errors where outlines deviate and the volume of the erroneous region is small compared to the total volume. Thus, we considered also distance-based metrics like the Hausdorff distance (HD) and the average symmetric surface distance (ASSD).
 
To detect anatomically implausible deformations we calculated the determinant of the Jacobian matrix for every point in a deformation vector field. The value of the determinant gives information about how the deformation changes the volume at a given point. A value of $>$ 1 indicates expansion, a value of 1 indicates no volume change, a value between 0 and 1 indicates shrinkage and a value $\leq$ 0 indicates image folding. For reasons of simplification we state only the mean of the determinant values $\pm$ standard deviation as well as the fraction of foldings (\textit{FoF}) per image.

Statistical analysis was performed with the Wilcoxon signed‐rank test. This test, which has for null hypothesis that the median group difference is zero, was chosen due to non‐normal distribution and heteroscedasticity of the data. In our experiments, the confidence alpha was set to 1\%.

\subsection{Experiments}

We evaluated the presented algorithm in two different ways for each 4D dataset. First, we were interested in the general performance of the presented algorithm by registering only the two images of each dataset depicting extreme phases (maximum inhale/exhale, end-systolic/diastolic). In a second step we evaluated the ability of the presented algorithm to conceive periodic deformations over depicted phases. 
The size of the patches sequentially processed by the network may have an influence on the results. A patch size too small could hamper the algorithm from capturing large deformations and has a large computational overhead by copying data to and from the GPU. A large patch size can prevent registration by running out of memory. To measure the impact of the patch size on the results we conducted the two evaluation ways mentioned before with patches of size 48$\times$48$\times$48 (\textit{Patch48}), 64$\times$64$\times$64 (\textit{Patch64}), 80$\times$80$\times$80 (\textit{Patch80}) and 96$\times$96$\times$96 voxels (\textit{Patch96}). With a patch size of 96 voxels we were running out of memory for some 4D cases, therefore larger patch sizes were not evaluated. 
In a last test setting the impact of the downsampling layers and whether the high computational effort for training each image patch till convergence can be reduced with a few shot approach \cite{Ferrante2018a} are analysed.

\subsubsection{3D registration evaluation}
For several datasets at hand landmarks or segmentations are only available for the two extreme phases. To analyse the registration accuracy and the ability of our algorithm to calculate a DVF and its inverse simultaneously just these two phases were employed. The registration of only two phases resembles a conventional 3D registration task and facilitates also a comparison to published results in the literature. The figures of merit mentioned in section \ref{sec:evaluation} were calculated for the DVF from exhale/diastolic phase to inhale/systolic phase and vice versa. To measure how well the inverse DVF is approximated by the presented algorithm we applied the two calculated DVF of a dataset to a landmark-set or a segmentation in sequence and compared the double deformed input to the original input. This procedure was carried out two times for each dataset. First by starting with the exhale/diastolic phase and in a second run by starting with the inhale/systolic phase. In an ideal case double deformed input and original input would be the same. To account for anatomically implausible deformations we calculated the Jacobi determinant for every DVF.

\subsubsection{4D registration evaluation}
When registering 4D datasets it is important that the algorithm can track moving structures over all phase images and takes care of the underlying motion pattern. In contrast to 3D registration the network has to combine multiple small deformations instead of calculating one large deformation. We investigated how the registration error changes from one phase image to another phase image, whether it increases or decreases compared to 3D registration when focusing on the extreme phases only and how well the periodic  motion pattern is reflected. This is done by calculating landmark distance, DSC, ASSD, HD for every single DVF from one phase image to its successor (provided landmarks or contour are available) as well as for combinations of successive DVF. The evaluation scheme for one phase image $i$ is defined as follows: 
\vspace{5pt}
\begin{enumerate}
\item set $j$ to $i+1$
\item deform landmarks/segmentations of a phase image $i$ with $T_i^{j}$ and compare the result to landmarks/segmentations of phase image $j$
\item if $j < N-1$ set $j$ to $j+1$ otherwise set $j$ to $0$
\item if $j$ is equal to $i$ terminate otherwise go to 2)
\end{enumerate}
\vspace{5pt}
This scheme is repeated for all phase images in a dataset $I^N$ and results in $N \times N$ comparisons provided that landmarks/segmentations are available for all phase images. If landmarks/segmentations are missing for a given phase image the comparison is skipped. In an ideal case the consecutively deformed landmarks/segmentations of a given phase image with all DVF of a dataset are equal to the original landmarks/segmentations of that given phase image.
Foldings were again detected by calculating the Jacobi determinant. 

\subsubsection{Downsampling And Few Shot Learning}

Calculating a dense displacement field by training the network till convergence for each image patch demands a lot of calculations and memory. To determine the gain of accuracy by the third layer of our network (which calculates a DVF without downsampling) we performed tests by using only the first two layers. 
In a second step, we analysed how the training on a few datasets influences computation time and registration error. We decided to do the second analysis only for the DirLab dataset as it is more homogeneous with respect to pathological changes compared to Popi and Sunnybrook. Due to the small number of datasets we conducted this test only in 3D and performed a leave-one-out cross validation. We performed the training for each downsampling step separately. To determine the number of epochs for the training phase we monitored the registration error of the test dataset during training. The optimal number of epochs for a test dataset was defined right before the error stagnates or increases again. To mitigate overfitting of the epochs parameter caused by this procedure we took the median number of the 10 optimal number of epochs and performed the training with the median again to gain the final models. A trained model was then used in two ways: first, it was applied to the test dataset without any further training; second, the model was fine tuned with the dataset to be registered for 50 epochs. Finally, we evaluated the impact of the knowledge obtained in the training phase on the registration process by stopping the one shot approach after 50 epochs and comparing it to the few shot approach fine tuned for 50 epochs.

\section{Results}

The first part of this section covers the analysis of the different input patch sizes. With the best performing patch size we then evaluated our algorithm for the registration of two 3D image datasets with a focus on accuracy, inverse calculation and comparison to other state-of-the-art algorithms. The third part deals with the evaluation of the 4D dataset registration results with an emphasis on accuracy and periodicity. In the last subsection results for the algorithm with only two layers and the few shot approach are given. 

\subsection{Patch size}

Table \ref{tab:patchSizeLayerRes} summarises the registration performance of our algorithm with different patch sizes per datasets. At a first glance the high computation time of our algorithm with a patch size of 48 voxles attracts attention. The higher the patch size the lower the computation time, which indicates that additional computations for copying data to and from the GPU and assembling the deformation field are an important factor. For the 3D test cases the worst results were given also with Patch48. Especially for the lung datasets a big difference to the other settings could be seen. A deeper inspection showed that the higher the initial registration error the higher is the difference to the better performing methods. Thus, we assume that a patch size of 48 voxels is already too small to gather enough information to account for large deformations. Therefore, Patch48 was disregarded in the remainder. The results for patch sizes of 64, 80 and 96 voxels look similar, which is also confirmed by a statistical analysis. Only for the datasets Sunnybrook3D, Popi4D and Dirlab4D the best performing algorithm yielded a statistically better performance compared to the others. The algorithm with a patch size of 80 voxels yielded best results in two out of these three test cases. Therefore, in further  analyses  and  comparisons, we report only the results of our algorithm with a patch size of 80 voxels.

\begin{table*}[ht!]
\centering
\caption{In this table image details of the datasets used or testing are listed in addition with the initial registration error and the results of the 3D baseline registration evaluation tests. For the lung datasets the average landmark distance (LD) $\pm$ standard deviation is given for maximum exhalation to inhalation phase registration and vice versa, average S\o{}rensen-Dice index (DSC), Hausdorff distance (HD) and average symmetric surface distance (ASSD) $\pm$ standard deviation of the contours for end-diastolic and end-systolic phase registrations can be seen for the cardiac MRI datasets.}
\label{tab:datasetDetails}
\begin{tabular}{lcccccccccccc}
\multicolumn{2}{l}{\textbf{\textsc{Lung Datasets}}} & & & \multicolumn{3}{c}{\textbf{Before Registration}} & \multicolumn{3}{c}{\textbf{\makecell{Inhale to Exhale}}} & \multicolumn{3}{c}{\textbf{\makecell{Exhale to Inhale}}} \\
\textbf{Dataset} & \textbf{Dimensions} & \textbf{\makecell{Time-\\bins}} & \textbf{Voxelsize (mm)} & \multicolumn{3}{c}{\textbf{\makecell{LD (mm)}}} & \multicolumn{3}{c}{\textbf{\makecell{LD (mm)}}} & \multicolumn{3}{c}{\textbf{\makecell{LD (mm)}}} \\
DirLab01  & 256 x 256 x 94 & 10  & 0.97 x 0.97 x 2.50 & \multicolumn{3}{c}{3.89 $\pm$ 2.78} & \multicolumn{3}{c}{1.26 $\pm$ 1.06} & \multicolumn{3}{c}{1.16 $\pm$ 0.65 }\\
DirLab02  & 256 x 256 x 112 & 10 & 1.16 x 1.16 x 2.50 & \multicolumn{3}{c}{4.34 $\pm$ 3.90} & \multicolumn{3}{c}{1.14 $\pm$ 0.71} & \multicolumn{3}{c}{1.13 $\pm$ 0.59}\\
DirLab03  & 256 x 256 x 104 & 10 & 1.15 x 1.15 x 2.50 & \multicolumn{3}{c}{6.94 $\pm$ 4.05} & \multicolumn{3}{c}{1.31 $\pm$ 0.85} & \multicolumn{3}{c}{1.33 $\pm$ 0.79}\\
DirLab04  & 256 x 256 x 99 & 10  & 1.13 x 1.13 x 2.50 & \multicolumn{3}{c}{9.83 $\pm$ 4.88} & \multicolumn{3}{c}{1.81 $\pm$ 1.78} & \multicolumn{3}{c}{1.87 $\pm$ 1.73}\\
DirLab05  & 256 x 256 x 106 & 10 & 1.10 x 1.10 x 2.50 & \multicolumn{3}{c}{7.47 $\pm$ 5.5} & \multicolumn{3}{c}{1.85 $\pm$ 1.81} & \multicolumn{3}{c}{1.75 $\pm$ 1.35}\\
DirLab06  & 512 x 512 x 128 & 10 & 0.97 x 0.97 x 2.50 & \multicolumn{3}{c}{10.89 $\pm$ 6.96} & \multicolumn{3}{c}{2.52 $\pm$ 4.90} & \multicolumn{3}{c}{2.09 $\pm$ 2.13}\\
DirLab07  & 512 x 512 x 136 & 10 & 0.97 x 0.97 x 2.50 & \multicolumn{3}{c}{11.02 $\pm$ 7.42} & \multicolumn{3}{c}{1.84 $\pm$ 1.66} & \multicolumn{3}{c}{1.98 $\pm$ 1.64}\\
DirLab08  & 512 x 512 x 128 & 10 & 0.97 x 0.97 x 2.50 & \multicolumn{3}{c}{14.99 $\pm$ 9.00} & \multicolumn{3}{c}{3.36 $\pm$ 4.75} & \multicolumn{3}{c}{3.57 $\pm$ 5.24}\\
DirLab09  & 512 x 512 x 128 & 10 & 0.97 x 0.97 x 2.50 & \multicolumn{3}{c}{7.92 $\pm$ 3.97} & \multicolumn{3}{c}{1.44 $\pm$ 0.84} & \multicolumn{3}{c}{1.50 $\pm$ 0.85}\\
DirLab10 & 512 x 512 x 120 & 10 & 0.97 x 0.97 x 2.50 & \multicolumn{3}{c}{7.30 $\pm$ 6.34} & \multicolumn{3}{c}{1.79 $\pm$ 2.52} & \multicolumn{3}{c}{1.80 $\pm$ 1.93}\\
Popi01 & 512 x 512 x 141 & 10 & 0.98 x 0.98 x 2.00 & \multicolumn{3}{c}{5.90 $\pm$ 2.73} & \multicolumn{3}{c}{1.04 $\pm$ 0.46} & \multicolumn{3}{c}{1.15 $\pm$ 0.84}\\
Popi02 & 512 x 512 x 169 & 10 & 0.98 x 0.98 x 2.00 & \multicolumn{3}{c}{14.04 $\pm$ 7.20} & \multicolumn{3}{c}{2.51 $\pm$ 2.89} & \multicolumn{3}{c}{2.90 $\pm$ 3.62}\\
Popi03 & 512 x 512 x 170 & 10 & 0.88 x 0.88 x 2.00 & \multicolumn{3}{c}{7.67 $\pm$ 5.05} & \multicolumn{3}{c}{1.43 $\pm$ 1.52} & \multicolumn{3}{c}{1.38 $\pm$ 1.56}\\
Popi04 & 512 x 512 x 187 & 10 & 0.78 x 0.78 x 2.00 & \multicolumn{3}{c}{7.33 $\pm$ 4.89} & \multicolumn{3}{c}{1.25 $\pm$ 1.91} & \multicolumn{3}{c}{1.10 $\pm$ 1.74}\\
Popi05 & 512 x 512 x 139 & 10 & 1.17 x 1.17 x 2.00 & \multicolumn{3}{c}{7.09 $\pm$ 5.08} & \multicolumn{3}{c}{1.32 $\pm$ 1.01} & \multicolumn{3}{c}{1.27 $\pm$ 0.92}\\
Popi06 & 512 x 512 x 161 & 10 & 1.17 x 1.17 x 2.00 & \multicolumn{3}{c}{6.68 $\pm$ 3.68} & \multicolumn{3}{c}{1.29 $\pm$ 1.01} & \multicolumn{3}{c}{1.25 $\pm$ 0.88}\\[1ex]
\multicolumn{4}{c}{} & \multicolumn{3}{c}{\textbf{8.33 $\pm$ 8.86}} & \multicolumn{3}{c}{\textbf{1.70 $\pm$ 2.26 }} & \multicolumn{3}{c}{\textbf{1.70 $\pm$ 2.04}} \\[1.5ex]

\multicolumn{2}{l}{\textbf{\textsc{Cardiac Datasets}}} &  & & \multicolumn{3}{c}{\textbf{Before Registration}} & \multicolumn{3}{c}{\textbf{\makecell{Systolic to Diastolic}}} & \multicolumn{3}{c}{\textbf{\makecell{Diastolic to Systolic}}}\\
\textbf{Dataset} & \textbf{Dimensions} & \textbf{\makecell{Time-\\bins}} & \textbf{Voxelsize (mm)} & \textbf{DSC} & \textbf{\makecell{HD \\ (mm)}} & \textbf{\makecell{ASSD \\ (mm)}} & \textbf{DSC} & \textbf{\makecell{HD \\ (mm)}} & \textbf{\makecell{ASSD \\ (mm)}} & \textbf{DSC} & \textbf{\makecell{HD \\ (mm)}} & \textbf{\makecell{ASSD \\ (mm)}}\\
SC-HF-I-1 & 256 x 256 x 12 & 20 & 1.37 x 1.37 x 10.00 &  0.84 & 14.73 & 2.05 & 0.93 & 8.08 & 1.40 & 0.90 & 12.00 & 1.58     \\
SC-HF-I-2 & 256 x 256 x 13 & 20 & 1.37 x 1.37 x 10.00 &  0.84 & 12.11 & 1.94 & 0.94 & 6.71 & 1.27 & 0.93 & 7.50 & 1.26 \\
SC-HF-I-4 & 256 x 256 x 10 & 20 & 1.29 x 1.29 x 8.00 &   0.87 & 13.05 & 1.85 & 0.96 & 6.00 & 0.86 & 0.94 & 6.71 & 1.04  \\
SC-HF-I-40 & 256 x 256 x 11 & 20 & 1.37 x 1.37 x 8.00 &  0.73 & 16.46 & 2.95 & 0.94 & 8.75 & 0.96 & 0.87 & 12.73 & 1.56\\
SC-HF-NI-3 & 256 x 256 x 12 & 20 & 1.37 x 1.37 x 8.00 &  0.90 & 11.68 & 1.25 & 0.96 & 4.74 & 0.82 & 0.94 & 7.65 &  0.99  \\
SC-HF-NI-4 & 256 x 256 x 11 & 20 & 1.37 x 1.37 x 8.00 &  0.86 & 9.40 & 1.96 & 0.93 & 7.50 & 1.21 & 0.89 & 15.00 & 1.80    \\
SC-HF-NI-34 & 256 x 256 x 13 & 20 & 1.37 x 1.37 x 10.00 & 0.78 & 12.30 & 2.17 & 0.92 & 10.50 & 1.41 & 0.89 & 13.08 & 1.57\\
SC-HF-NI-36 & 256 x 256 x 10 & 20 & 1.21 x 1.21 x 10.00 & 0.78 & 12.30 & 2.17 & 0.95 & 20.00 & 0.59 & 0.93 & 11.11 & 0.84 \\
SC-HYP-1 & 256 x 256 x 9 & 20 & 1.37 x 1.37 x 10.00 & 0.53 & 20.05 & 5.13 & 0.84 & 15.00 & 2.48 & 0.68 & 16.50 & 3.02     \\
SC-HYP-3 & 256 x 256 x 8 & 20 & 1.37 x 1.37 x 10.00 & 0.32 & 19.91 & 8.94 & 0.89 & 10.50 & 1.24 & 0.71 & 13.50 & 2.17 \\
SC-HYP-38 & 256 x 256 x 12 & 20 & 1.37 x 1.37 x 10.00 & 0.71 & 12.60 & 2.47 & 0.88 & 7.79 & 1.68 & 0.61 & 16.84 & 3.02 \\
SC-HYP-40 & 256 x 256 x 16 & 20 & 1.37 x 1.37 x 8.00 &  0.55 & 20.14 & 6.44 & 0.88 & 6.71 & 1.55 & 0.78 & 11.52 & 1.95 \\
SC-N-2 & 256 x 256 x 9 & 20 & 1.37 x 1.37 x 8.00 & 0.54 & 21.96 & 6.98 & 0.93 & 6.00 & 0.90 & 0.78 & 10.06 & 1.74   \\
SC-N-3 & 256 x 256 x 20 & 20 & 1.37 x 1.37 x 8.00 & 0.62 & 20.62 & 4.22 & 0.76 & 21.53 & 3.33 & 0.67 & 25.01 & 3.75  \\
SC-N-40 & 256 x 256 x 12 & 20 & 1.25 x 1.25 x 10.00 & 0.56 & 17.29 & 4.72 & 0.80 & 15.37 & 2.63 & 0.66 & 17.62 & 3.02  \\[1ex]
\multicolumn{4}{c}{} & \textbf{0.70} & \textbf{15.64} & \textbf{3.68} & \textbf{0.90} & \textbf{10.35} & \textbf{1.49} & \textbf{0.81} & \textbf{13.12} & \textbf{1.95} \\
\multicolumn{4}{c}{} & \textbf{$\pm$} & \textbf{$\pm$} & \textbf{$\pm$} & \textbf{$\pm$} & \textbf{$\pm$} & \textbf{$\pm$} & \textbf{$\pm$} & \textbf{$\pm$} & \textbf{$\pm$} \\
\multicolumn{4}{c}{} & \textbf{0.17} & \textbf{4.07} & \textbf{2.30} & \textbf{0.06} & \textbf{5.23} & \textbf{0.77} & \textbf{0.12} & \textbf{4.71} & \textbf{0.87} \\
\end{tabular}
\end{table*}

\begin{table*}[ht!]
\centering
\caption{This table compares the registration performance of our algorithm with four different patch size settings ranging from 48$\times$48$\times$48 voels to 96$\times$96$\times$96 voxels. The last column shows the computation time averaged over 3D and 4D datasets. \mbox{* Landmark distance;} ** S\o{}rensen-Dice index}
\label{tab:patchSizeLayerRes}
\begin{tabular}{lccccccc}
\textbf{Method} & \textbf{\makecell{Dirlab3D \\ (mm)*}} & \textbf{\makecell{Popi3D \\(mm)*}} & \textbf{\makecell{Sunnybrook3D \\ (DSC)**}} & \textbf{\makecell{Dirlab4D \\ (mm)**}} & \textbf{\makecell{Popi4D \\ (mm)*}} & \textbf{\makecell{Sunnybrook4D \\ (DSC)**}} & \textbf{\makecell{Computation Time \\ (min)}}\\
Patch48 & 1.92 $\pm$ 2.35 & 1.77 $\pm$ 2.03 & 0.83 $\pm$ 0.12 & 1.99 $\pm$ 1.78 & 1.75 $\pm$ 1.36 & 0.87 $\pm$ 0.10 & 44 $\pm$ 50\\
Patch64 & 1.77 $\pm$ 2.07 & 1.50 $\pm$ 1.67 & 0.85 $\pm$ 0.11 & 1.91 $\pm$ 1.58 & 1.73 $\pm$ 1.24 & 0.89 $\pm$ 0.09 & 31 $\pm$ 33 \\
Patch80 & 1.83 $\pm$ 2.35 & 1.49 $\pm$ 1.59 & 0.86 $\pm$ 0.10 & 1.95 $\pm$ 1.64 & 1.71 $\pm$ 1.32 & 0.87 $\pm$ 0.13 & 28 $\pm$ 29\\
Patch96 & 1.86 $\pm$ 2.08 & 1.44 $\pm$ 1.50 & 0.84 $\pm$ 0.12 & 1.99 $\pm$ 1.57 & 1.77 $\pm$ 1.30 & 0.87 $\pm$ 0.15 & 26 $\pm$ 26\\
\end{tabular}
\end{table*}

\subsection{3D registration evaluation}

In table \ref{tab:datasetDetails} the results for the 3D registration evaluation can be seen. The average landmark distance between maximum inhale and exhale phase of the lung datasets could be reduced from an initial value of 8.33 mm to 1.70 mm for both, maximum inspiration to expiration phase and vice versa. DSC could be increased to a value of 0.86 and HD and ASSD could be decreased to values of 11.73 and 1.72, respectively with DVF calculated between end-diastolic and end-systolic phase of the cardiac cine MRI datasets. The analysis of the ability to calculate the inverse transformation showed an average landmark distance of 0.89 $\pm$ 0.42 mm comparing the original landmark set with the two times deformed landmark sets of the lung datasets. The same analysis for the Sunnybrook datasets showed DSC, HD and ASSD values of 0.97 $\pm$ 0.02, 6.17 $\pm$ 4.27 mm and 0.36 $\pm$ 0.25 mm, respectively. The large difference of HD and ASSD can be explained with the big slice thickness of several millimetres.

\begin{figure}[htb]
     \begin{center}
     \mbox{
      \shortstack{
        \includegraphics[width=0.45\linewidth]{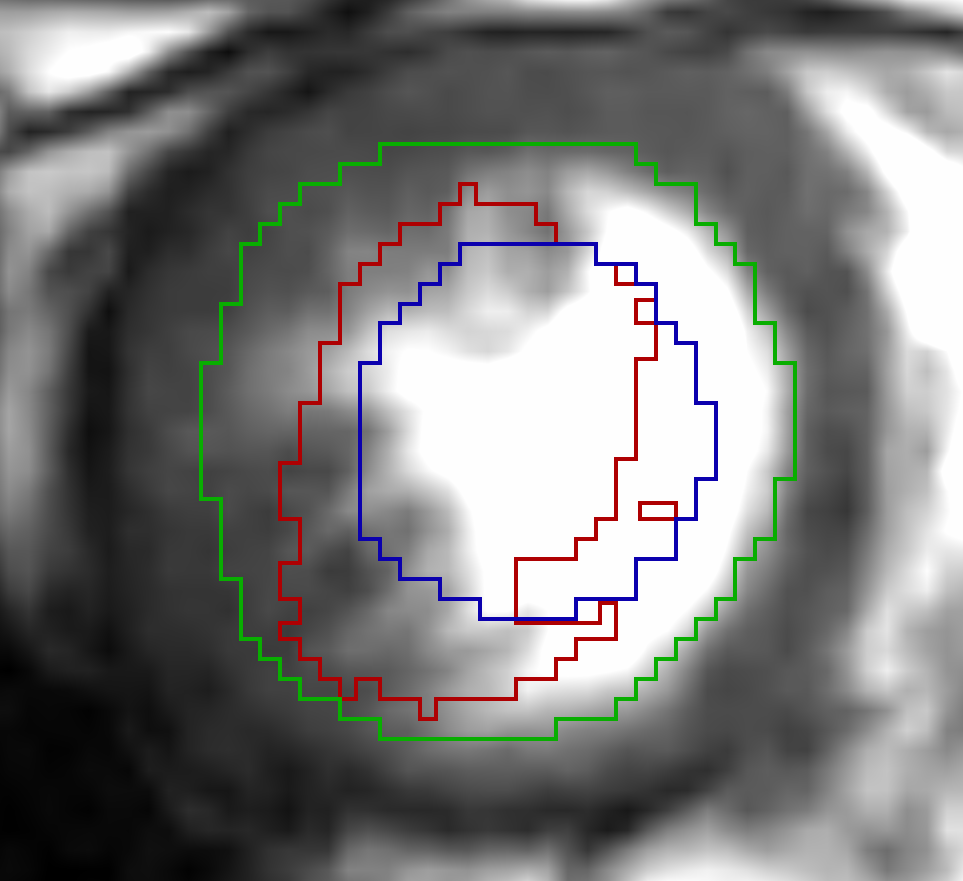} \\
        a) x,y-plane of SC-HYP-38
        }
        \hspace{-1.5 mm}
         
      \shortstack{     
        \includegraphics[width=0.45\linewidth]{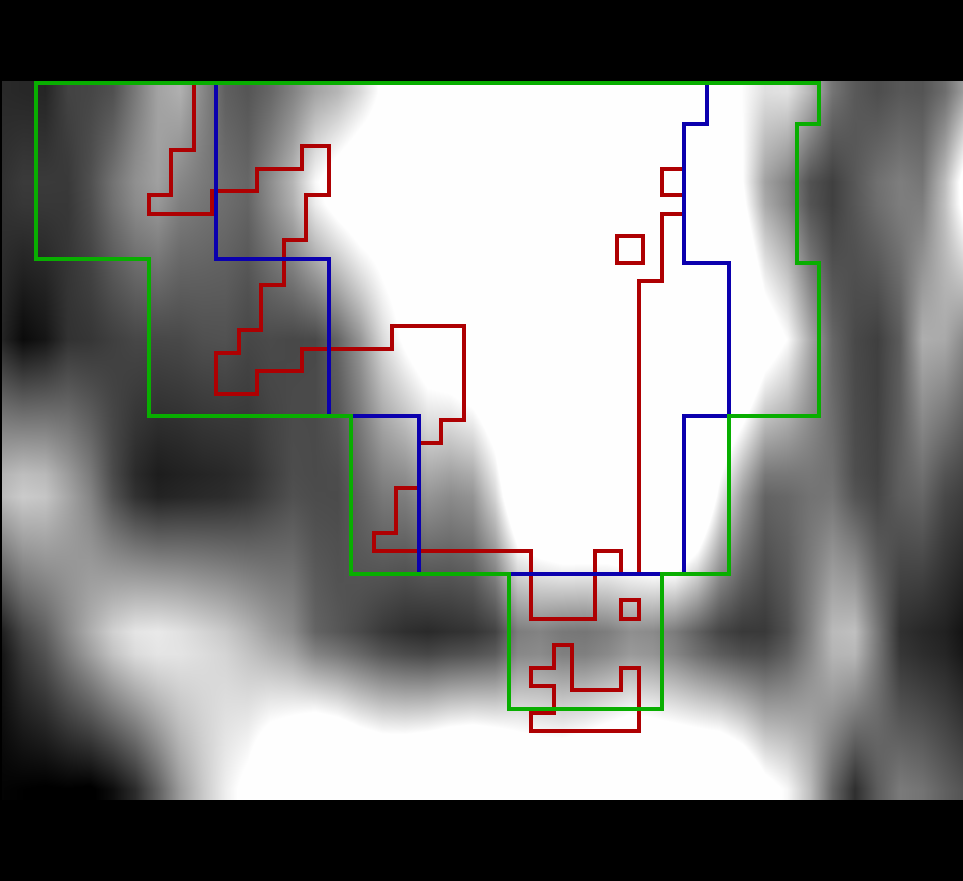} \\
        b) SC-HYP-38 along z-axis
        }
               
        }               

        \caption{In this figure the result for registering extreme systolic to diastolic phase for case SC-HYP-38 is depicted. The spikes in z-direction of the calculated contour (red) in b) and the closeness of the input contour (green) to tissue with a lower motion amplitude are an indicator that a reason for the poor registration result lies in the resampling along the z-axis and the nearest neighbour interpolation.}
\label{fig:Sunny10Analysis}
\end{center}        
\end{figure}

For the cardiac datasets the DVF for mapping the systolic phase to diastolic phase showed better results (DSC: 0.90, HD: 10.35 mm, ASSD: 1.49 mm) compared to the registration from diastolic to systolic phase (DSC: 0.81, HD: 13.12 mm, ASSD: 1.95 mm). We attribute this direction-dependent behaviour to the resampling of the Sunnybrook datasets. In the diastolic images the border of the LV segmentation is in the area of a high gradient in the deformation field where small and large deformations meet. Due to the nearest neighbour interpolation used for resampling the segmentation masks, the mask for interpolated slices is either too small or too large with respect to the linearly interpolated image slices. An example is shown in Fig. \ref{fig:Sunny10Analysis}. However, to exclude an impact of the networks design we repeated the registration with switched input channels. The repeated experiments showed similar results and confirm the more accurate evaluation results from systolic to diastolic phase. Another point that contradicts a methodological failure is that for the lung datasets the better performance was distributed equally.

Table \ref{tab:resultComparison} gives results of our proposed algorithm alongside those reported in the literature for DL based and conventional registration algorithms. Summarised, our algorithm is able to keep up with the results in the literature. One has to note that in the literature the results are stated only for one registration direction. As it was not always clear which registration direction the authors reported, we averaged our results over both directions.

\begin{table*}[ht!]
\newcolumntype{C}[1]{>{\centering}p{#1}}
\newcolumntype{L}[1]{>{\raggedright}p{#1}}
\centering
\caption{Comparison of registration results between our proposed registration algorithm and those reported in the literature. The algorithms by De Voss et al. \cite{Vos2019}, Eppenhof et al. \cite{Eppenhof2018, eppenhofSPIE} use deep learning based methods, the algorithms by Vishnevskiy et al. \cite{Vishnevskiy2017}, Vandemeulebroucke et al. \cite{vandemeulebroucke2011spatiotemporal} and Metz et al. \cite{Metz2011} are conventional registration algorithms. For the DirLab and Popi datasets the table shows the average landmark distances in mm, for the Sunnybrook dataset the average S\o{}rensen-Dice index is given. In case of the 3D evaluation the presented algorithm calculates the vector fields from one image to another and the inverse at the same time. The results of our algorithm shown in this table are averaged over both deformations. In the literature the results are reported only for one direction. The 4D evaluation shows the average landmark distance between consecutive phase images. The Sunnybrook datasets contain only segmentations in the extreme phases. Therefore, no evaluation between consecutive phase images was possible.}
\label{tab:resultComparison}
\begin{tabular}{lccccc|cc}
\multicolumn{6}{c|}{\textbf{\textsc{3D Evaluation}}} & \multicolumn{2}{c}{\textbf{\textsc{4D Evaluation}}}\\
\textbf{Dataset} & \textbf{Proposed} & \textbf{\makecell{De Voss \\ 2019 \cite{Vos2019}}} & \textbf{\makecell{Eppenhof \\ 2018 \cite{Eppenhof2018}}} & \textbf{\makecell{Eppenhof \\  2018 \cite{eppenhofSPIE}}} & \textbf{\makecell{Vishnevskiy \\ 2017 \cite{Vishnevskiy2017}}} & \textbf{Proposed} & \textbf{\makecell{Metz \\ 2011 \cite{Metz2011}}}\\
DirLab01 & 1.21 $\pm$ 0.88 & 1.27 $\pm$ 1.16 & - & 1.65 0.89 & 0.76 $\pm$ - & 1.05 $\pm$ 0.87 & 1.04 $\pm$ 0.86\\
DirLab02 & 1.13 $\pm$ 0.65 & 1.20 $\pm$ 1.12 & 1.24 $\pm$ 0.61 & 2.26 $\pm$ 1.16 & 0.77 $\pm$ - & 1.05 $\pm$ 0.81 & 0.97 $\pm$ 0.85\\
DirLab03 & 1.32 $\pm$ 0.82 & 1.48 $\pm$ 1.26 & - & 3.15 $\pm$ 1.63 & 0.90 $\pm$ - & 1.39 $\pm$ 0.82 & 1.37 $\pm$ 0.87 \\
DirLab04 & 1.84 $\pm$ 1.76 & 2.09 $\pm$ 1.93 & 1.70 $\pm$ 1.00 & 4.24 $\pm$ 2.69 & 1.24 $\pm$ - & 1.59 $\pm$ 1.11 & 1.54 $\pm$ 1.12 \\
DirLab05 & 1.80 $\pm$ 1.60 & 1.95 $\pm$ 2.10 & - & 3.52 $\pm$ 2.23 & 1.12 $\pm$ - & 1.61 $\pm$ 1.50 & 1.55 $\pm$ 1.47 \\
DirLab06 & 2.30 $\pm$ 3.78 & 5.16 $\pm$ 7.09 & - & 3.19 $\pm$ 1.50 & 0.80 $\pm$ - & 1.88 $\pm$ 1.88 & 1.84 $\pm$ 1.71\\
DirLab07 & 1.91 $\pm$ 1.65 & 3.05 $\pm$ 3.01 & - & 4.25 $\pm$ 2.08 & 0.80 $\pm$ - & 1.64 $\pm$ 1.25 & 1.72 $\pm$ 1.38 \\
DirLab08 & 3.47 $\pm$ 5.00 & 6.48 $\pm$ 5.37 & - & 9.03 $\pm$ 5.08 & 1.34 $\pm$ - & 2.07 $\pm$ 1.97 & 2.61 $\pm$ 2.63 \\
DirLab09 & 1.47 $\pm$ 0.85 & 2.10 $\pm$ 1.66 & 1.61 $\pm$ 0.82 & 3.85 $\pm$ 1.86 & 0.92 $\pm$ - & 1.41 $\pm$ 0.87 & 1.33 $\pm$ 0.94 \\
DirLab10 & 1.79 $\pm$ 2.24 & 2.09 $\pm$ 2.24 & - & 5.07 $\pm$ 2.31 & 0.82 $\pm$ - & 1.69 $\pm$ 1.38 & 1.66 $\pm$ 1.41 \\
\textbf{Average} & \textbf{1.83 $\pm$ 2.35} & \textbf{2.64 $\pm$ 4.32} & \textbf{1.52 $\pm$ 0.85} & \textbf{4.02 $\pm$ 3.08} & \textbf{0.95 $\pm$ -} & \textbf{1.54 $\pm$ 1.31} & \textbf{1.56 $\pm$ 1.42} \\[1ex]
\textbf{Dataset} & \textbf{Proposed} & \textbf{\makecell{Vandemeulebroucke \\ 2011 \cite{vandemeulebroucke2011spatiotemporal}}} & & & & \textbf{Proposed} & \textbf{\makecell{Metz \\ 2011 \cite{Metz2011}}}\\
Popi01 & 1.09 $\pm$ 0.68 & 0.96 $\pm$ 0.57 & & & & 1.16 $\pm$ 0.92 & 1.01 $\pm$ 0.70\\
Popi02 & 2.71 $\pm$ 3.28 & 1.56 $\pm$ 1.34 & & & & 1.47 $\pm$ 1.05 & 1.39 $\pm$ 1.15\\
Popi03 & 1.40 $\pm$ 1.54 & 1.53 $\pm$ 1.70 & & & & 1.09 $\pm$ 0.70 & 0.95 $\pm$ 0.67\\
Popi04 & 1.17 $\pm$ 1.83 & 1.96 $\pm$ 2.92 & & & & - & - \\
Popi05 & 1.30 $\pm$ 0.97 & 1.48 $\pm$ 1.39 & & & & - & - \\
Popi06 & 1.27 $\pm$ 0.95 & 1.25 $\pm$ 0.95 & & & & - & - \\
\textbf{Average} & \textbf{1.49 $\pm$ 1.59} & \textbf{1.46 $\pm$ 1.65} & & & & \textbf{1.24 $\pm$ 0.90} & \textbf{1.11 $\pm$ 0.87}\\[1ex]
\textbf{Dataset} & \textbf{Proposed} & \textbf{\makecell{De Voss \\ 2019 \cite{Vos2019}}} & & & & & \\
Sunnybrook & 0.86 $\pm$ 0.10  & 0.89 $\pm$ 0.14 & & & & &
\end{tabular}
\end{table*}

Analysis of the DVF with respect to anatomically implausible deformations showed that the FoF per image dataset was on average 0.25 \% of the calculated deformation vectors. The average Jacobi determinant value was 1.00 $\pm$ 0.32. A deeper inspection of the Jacobi determinant values revealed that folding occurs mainly in the area of large deformations close to the image or patch border where tissue can move out of or into the field of view.

\subsection{4D Evaluation}

\begin{figure}[htb]
     \begin{center}
     \mbox{
      \shortstack{
        \includegraphics[width=0.50\linewidth]{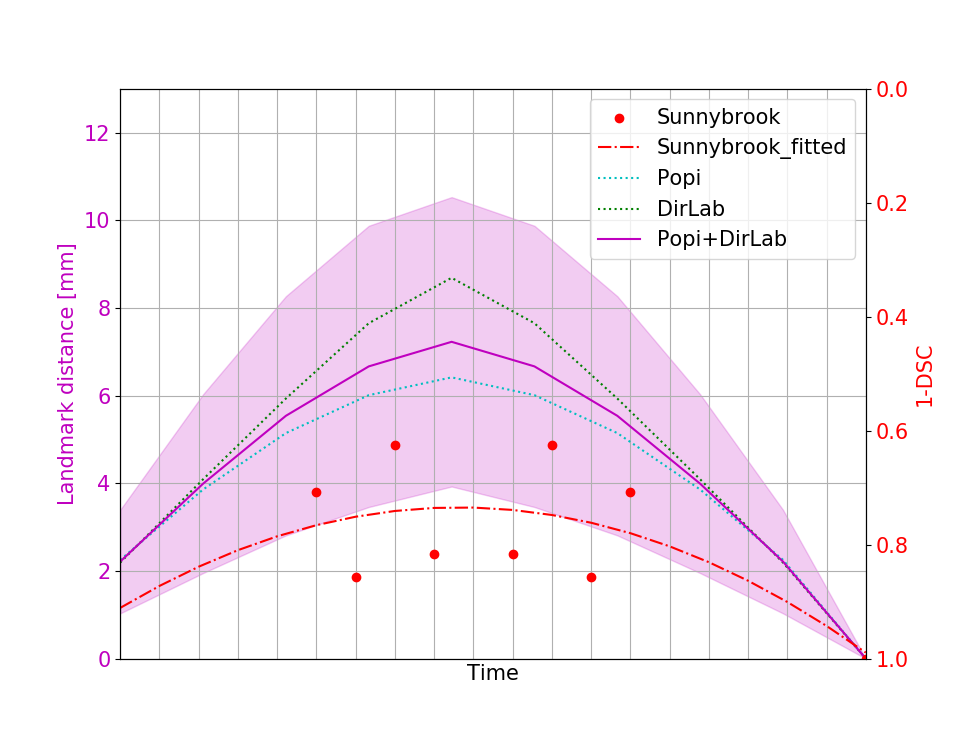} \\
        a) before registration
        }

      \shortstack{     
        \includegraphics[width=0.50\linewidth]{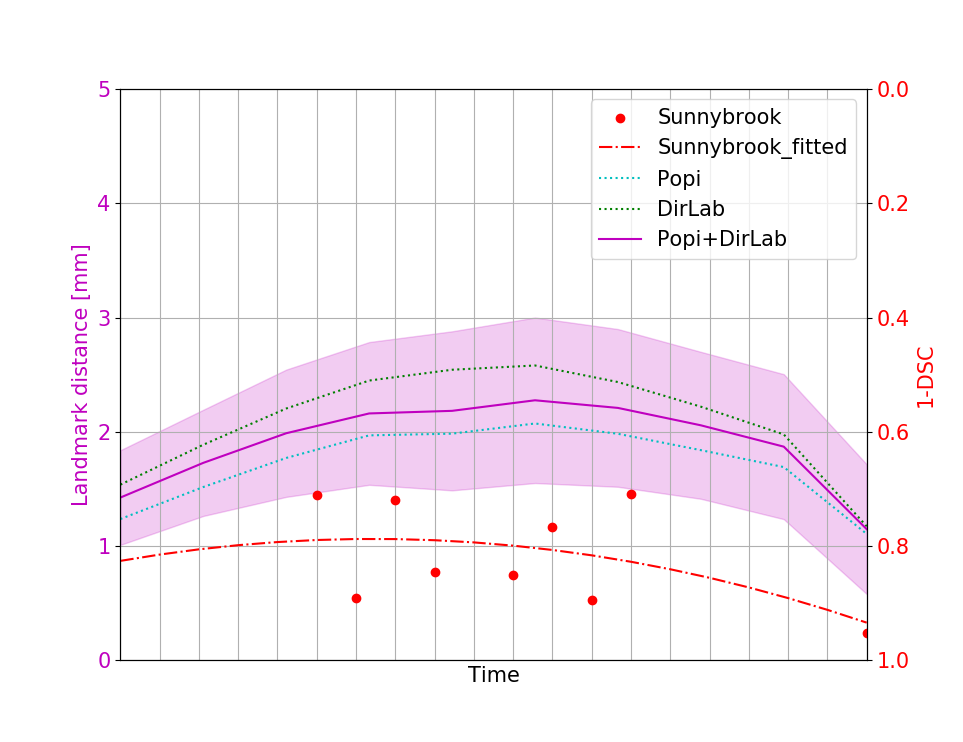} \\
        b) after registration
        }    
        }     
     \caption{In this figure the development of the registration error over time is plotted for the different 4D datasets before a) and after registration b). In the beginning the error is low, then due to large deformations and the combination of deformation vector fields the error increases towards the middle of a time series until it decreases again because of the periodicity constraint. For the cardiac Sunnybrook datasets only segmentations for the 2 extreme phases are available, therefore we interpolated the remaining values (red line). The magenta filled area indicates the average registration error for the Popi and DirLab datasets $\pm$ standard deviation.}
     \label{fig:cycleConsistency}
     \end{center}
     
\end{figure}

The initial average landmark distance between consecutive phase images was 2.25 $\pm$ 1.55 mm for the Popi datasets and 2.19 $\pm$ 2.02 mm for the DirLab datasets. The proposed registration algorithm could decrease the average distance to 1.24 $\pm$ 0.90 mm and 1.54 $\pm$ 1.31 mm, respectively. For a comparison with existing 4D registration methods we applied the publicly available method by \cite{Metz2011} to the respiratory datasets and achieved average registration errors similar to the ones of our method: 1.56 $\pm$ 1.42 mm for the DirLab and 1.11 $\pm$ 0.87 mm for the Popi datasets.
The construction of trajectories for voxels requires the combination of all DVF of a 4D dataset. As a consequence the registration error at the start of a trajectory gets propagated over time and increases or decreases with the combination of the different DVF. In Fig. \ref{fig:cycleConsistency} the error evolution is visualised for the three test datasets. At the beginning the mean registration error is 1.54 mm for the DirLab datasets and 1.24 mm for the Popi datasets. The error then increases to a mean of 2.58 mm for DirLab and 2.07 mm for Popi in the middle of the periodic motion. In the end of the breathing cycle the error falls to an error around 1 mm for both datasets. A similar error propagation pattern could be measured for the Sunnybrook datasets. For the Sunnybrook datasets segmentations are only available for the extreme phases, which allows only four comparisons per dataset. When starting with the end-diastolic phase a comparison in the middle of the cyclic motion to the end-systolic segmentation and in the end of the cycle motion to the original end-diastolic segmentation is possible. The same applies when starting with the end-systolic segmentation. In Fig. \ref{fig:cycleConsistency} the red dots indicate the measured DSC overlap, the red dotted line was fitted to the dots and estimates the DSC overlap for phase images without a segmentation available. 
Compared to a direct registration of the maximum exhale and inhale phases the registration error increased on average by 0.48 mm with the combination of the DVF. The average registration error for the extreme phases in the Popi dataset increased from 1.46 $\pm$ 1.59 mm to 1.98 $\pm$ 1.56 mm when we registered the whole 4D dataset instead of the two extreme phases solely. The same analysis reported an increase from 1.83 $\pm$ 2.35 mm to 2.54 $\pm$ 2.01 mm for the DirLab images and and decrease in the DSC from 0.86 $\pm$ 0.10 to 0.79 $\pm$ 0.14 for the Sunnybrook data.
The DVF of the 4D evaluation showed a FoF of 0.02 \% and an average Jacobi determinant of 1.00 $\pm$ 0.06.

\subsection{Downsampling And Few Shot Learning}

When we used only the first two layers of our network for registration the performance decreased slightly but the gain in computation time was huge. The average registration error for registering 3D respiratory images was 1.84 $\pm$ 1.88 mm compared to 1.70 $\pm$ 2.15 mm with all 3 layers. The computation time for two 3D images was on average 4 minutes, for 4D datasets the average time needed for registration increased to 8 minutes. The peak in error propagation increased to an average value of 2.32 $\pm$ 1.55 mm, 3.08 $\pm$ 2.27 mm and 0.75 $\pm$ 0.14 DSC for the Popi, DirLab and Sunnybrook datasets, respectively.
The few shot analysis on the DirLab dataset worked only for a setting with two downsampling layers. Training the third layer always resulted in an increase of the registration error for the dataset to be registered. Which is an indication that the network could not learn a general representation of detailed deformation from the limited amount of training data. Downsampling layer 1 and 2 were trained for 200 and 250 epochs, respectively. Applying the trained 2 layer models to the test data resulted in an registration error of 4.43 $\pm$ 2.94 mm, but by fine tuning the model for 50 epochs the registration error could be decreased to 2.16 $\pm$ 2.20 mm, which is comparable to the registration error of the one shot approach with 2.00 $\pm$ 2.03 mm for the DirLab dataset and 2 layers. With the few shot approach the computation time could be reduced by a factor of 4 to 1 minute. 

Stopping the one shot approach after 50 epochs resulted in an average landmark distance of 2.64 $\pm$ 2.21 mm. To achieve the same accuracy as the few shot approach another 30 epochs were needed. These numbers illustrate nicely the significant impact of the knowledge gained during the training phase.



\begin{figure*}[htb]
     \begin{center}
     \mbox{
      \shortstack{
        \includegraphics[width=0.30\linewidth]{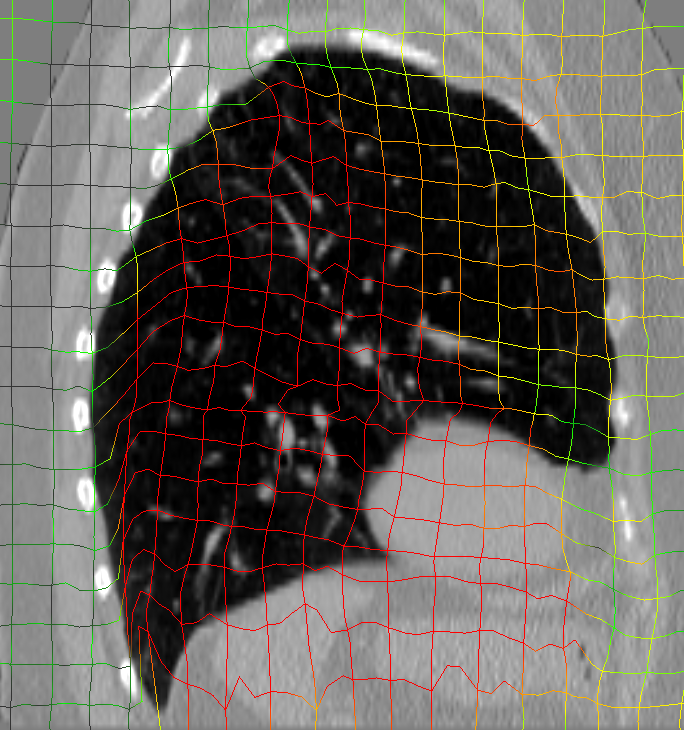} \\
        a) DirLab08 deformation grid \\ for inhale to exhale registration
        }
        \hspace{-1.5 mm}
         
      \shortstack{     
        \includegraphics[width=0.30\linewidth]{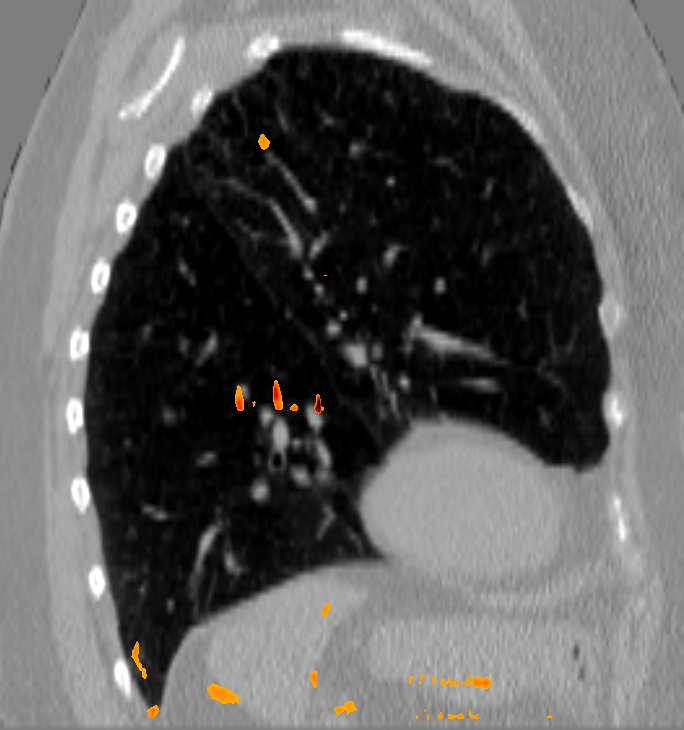} \\
        b) DirLab08 with overlayed image foldings \\ (Jacobi determinant $<$ 0)
        }
        \hspace{-1.5 mm}

        \shortstack{
        \includegraphics[width=0.30\linewidth]{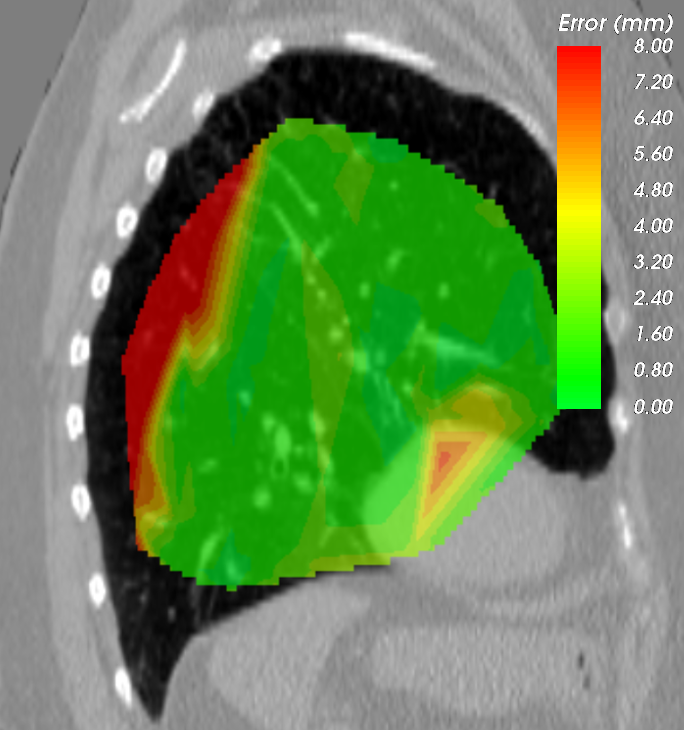} \\
        c) DirLab08 registration error \ \vspace{3.5 mm}
        }
               
        }

     \mbox{
      \shortstack{
        \includegraphics[width=0.30\linewidth]{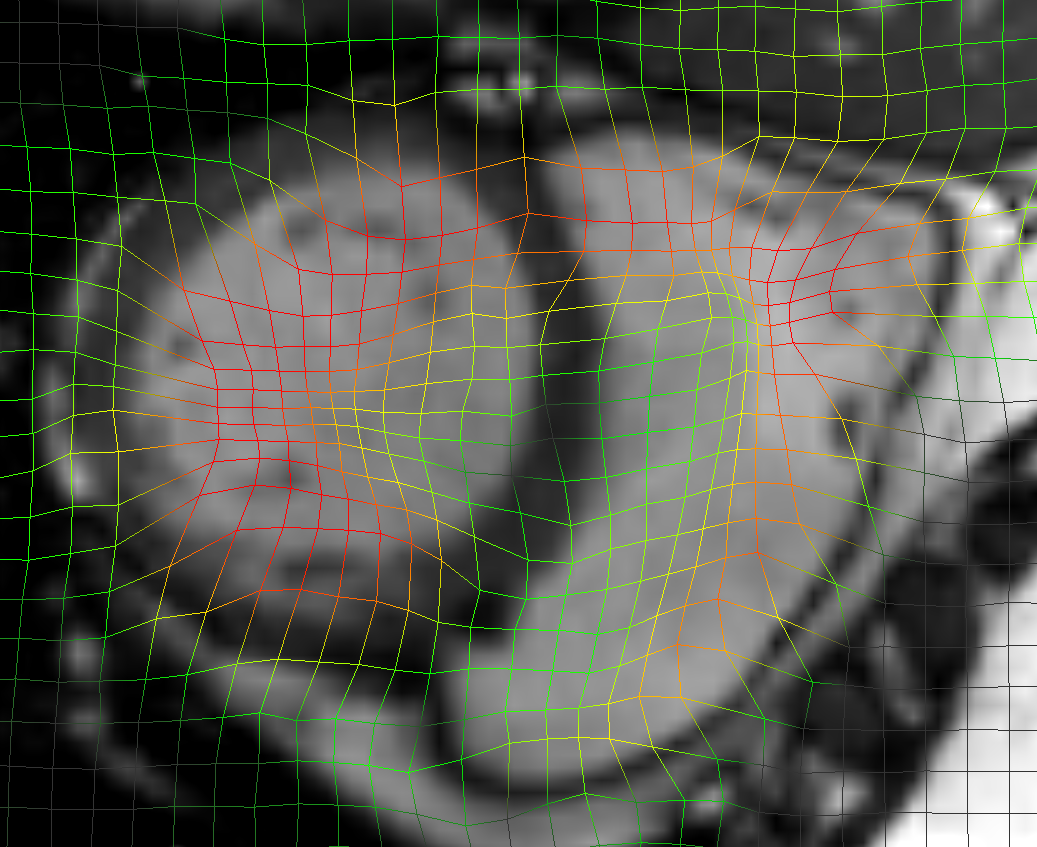} \\
        d) SC-N-3 deformation grid \\ for diastolic to systolic registration
        }
        \hspace{-1.5 mm}
         
      \shortstack{     
        \includegraphics[width=0.30\linewidth]{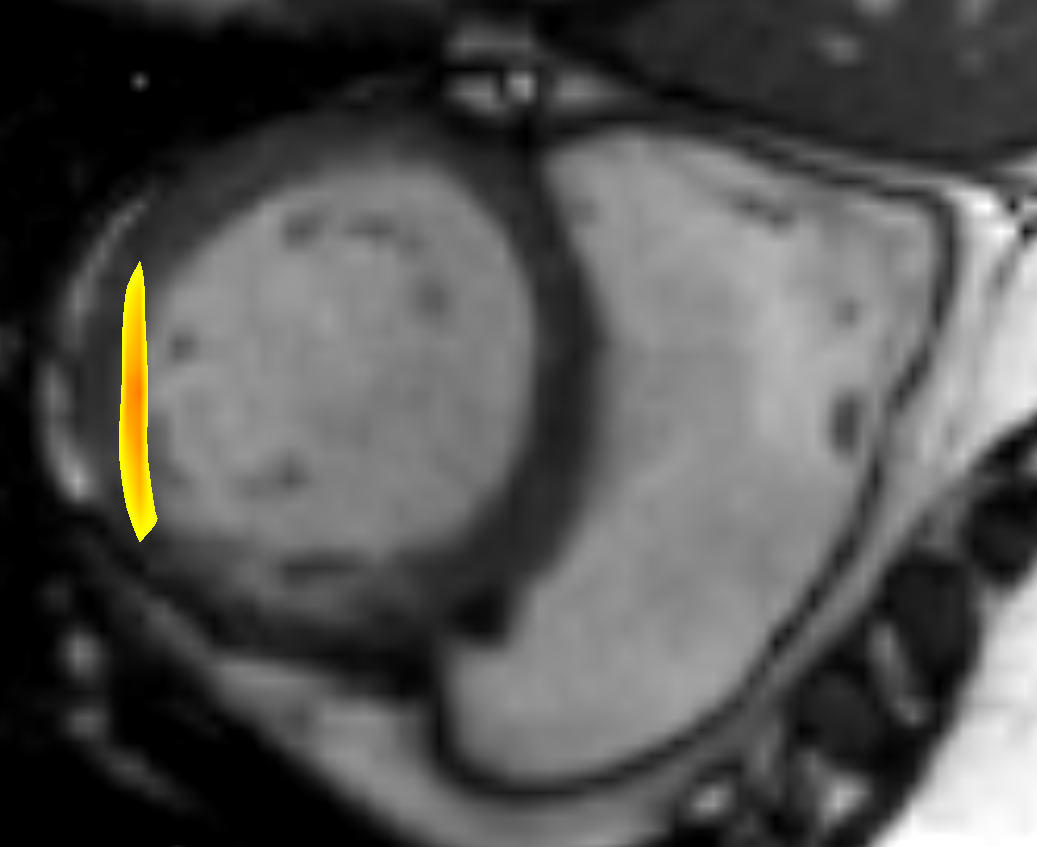} \\
        e) SC-N-3 with overlayed image foldings \\ (Jacobi determinant $<$ 0)
        }
        \hspace{-1.5 mm}

        \shortstack{
        \includegraphics[width=0.30\linewidth]{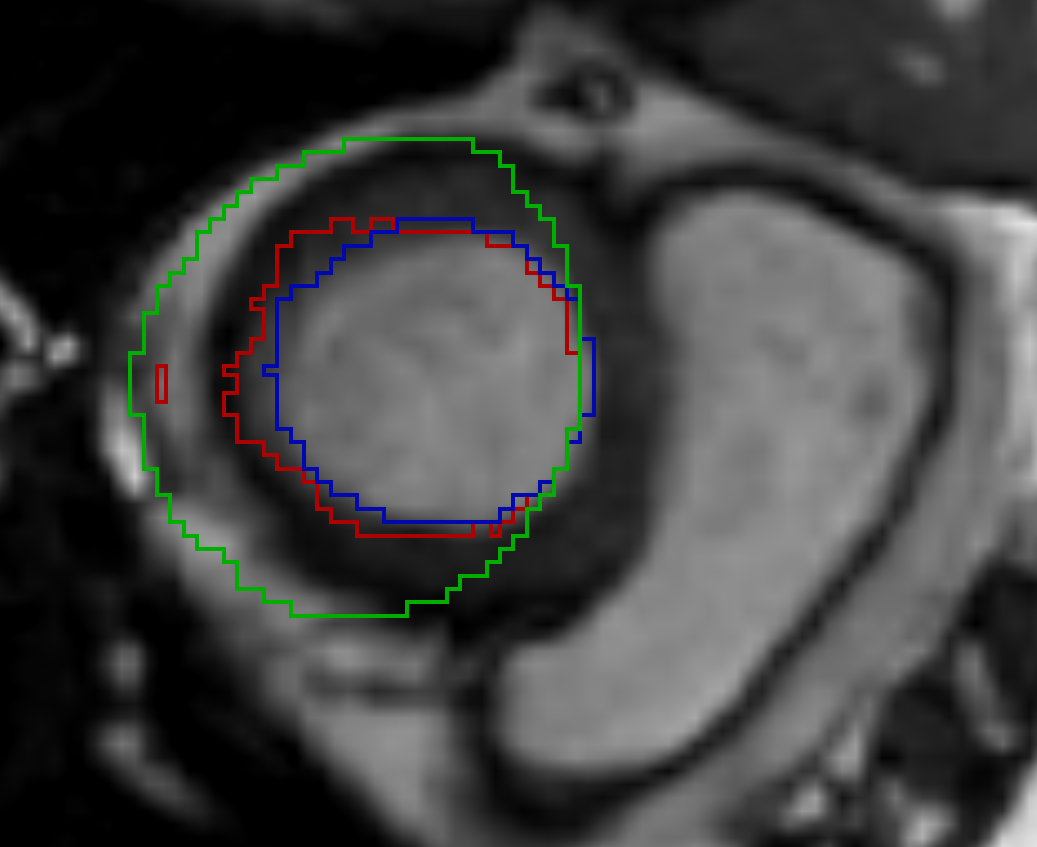} \\
        f) SC-N-3 contours (green: input, \\ blue: target, red: result)
        }

        }                     

        \caption{This figures show an analysis for two test cases that resulted in the worst results for 3D registration evaluation. The first row shows the extreme inhale phase of dataset Dirlab08 with overlayed deformation grid (a), Jacobi determinant (b) and registration error (c). The patient had a very strong sheering motion along the ribcage in caudal direction which could not be fully captured by our network. The second row shows the extreme diastolic phase for the dataset SC-N-3 with overlayed deformation grid (d) and Jacobi determinant (e) and the extreme systolic phase with contours of the left ventricle (f). }
\label{fig:ResultAnalysis}
\end{center}        
\end{figure*}

\section{Discussion}

We proposed a new method to combine DL and classical medical image registration techniques to track organ motion on 3D and 4D medical image datasets. Compared to other methods our approach offers three main benefits. First, no training data are needed to achieve good results. Second, when the algorithm is applied to register two 3D datasets, it calculates the DVF and an approximation of its inverse simultaneously. Third, we demonstrated that our algorithm achieves a good performance independent of organ site or modality. 

We showed that our algorithm can calculate the motion between two 3D images and adequately approximate the inverse. Also the periodic motion of the beating heart or the respiratory cycle could be reflected by our algorithm. When we compared the direct registration of the two extreme phases of a dataset with the registration of the whole 4D dataset, an increase of the registration error at the extreme phases could be seen. In our work we decided to register each image to its directly adjacent timely neighbour. An approach that could probably alleviate this problem would be to register each phase image to a reference image. The reasons why we decided against the registration with a reference image are that our approach is more intuitive and directly calculates a trajectory for each voxel. Moreover we made the experience in the beginning of the work that the network is better in registering images with small deformations than in registering images with large deformations.

By evaluating the proposed algorithm with multiple 3D and 4D publicly available datasets we were able to show that the benefits of a DL architecture can be used without training data to achieve results as good as those reported in the literature with training data available. However, the results showed also that our algorithm is outperformed in 3D registration tasks by conventional image registration algorithms. One has to take into account that the comparison with conventional algorithms is limited due to the usage of different loss funtions (e.g. \cite{Vishnevskiy2017} makes use of motion specific parameters and weights), image features and optimisation techniques. The goal of this work is not prove the superiority of one image registration algorithm but to show a way how deep learning methods can be used in an image registration problem with no or only a small amount of training datasets.

The approach we used in this work is unsupervised but before a registration problem can be solved the influence of the different parts of the objective function needs to be tuned. This task requires expert knowledge and has a big impact on the accuracy. Due to simplicity we used one parameter set for all datasets. Although the general performance of our algorithm is very promising we made the experience that a different parameter setting would have increased the performance for some datasets. Here supervised methods definitely have an advantage.

One drawback of the presented registration algorithm is the long computation time compared to DL based registration methods with an extra training phase. Our few shot analysis showed that with a small training dataset and a short fine tuning step this drawback can be mitigated significantly with only a slight reduction of the registration performance. However, with the 9 datasets used for training the few shot approach worked out only with downsampled images. For learning small detailed deformations with the original image data more training data are needed. 

On grounds of our analysis we see two applications for our algorithm: When the training of a network is not possible due to a lack of datasets or very heterogeneous datasets, which is often the case in small mono-centric studies in the field of radiotherapy that investigate a specific organ motion for treatment planning. Such studies usually have a low patient number as the acquisition of 4D images is time consuming, costly and implicates a higher radiation exposure in case of 4D CT. A second application scenario would be e.g. in the early stage of a study with a larger patient cohort. Datasets could be registered with our one shot approach until enough datasets are available to allow for an extra training phase. Subsequently the same architecture used for one shot registration could be trained on the existing datasets to enable a faster registration of upcoming datasets with similar accuracy. In a lifelong learning fashion \cite{Parisi2019}, the network would learn over time and increase its performance with additional datasets. 

In our future work a focus will be on an easier optimisation of the objective function parameters. Especially in the case of a deep architecture the tuning of the meta parameters is very time consuming and unintuitive as it is hard to comprehend how a change in the parameters influences the network weights and the registration performance. Depending on the registration task at hand, different parameters are needed. For datasets with a large amount of shearing motion a lower weighting of the smoothness term could increase the registration quality. On the other hand for images with a homogeneous motion but with a lot of noise (like in some ultrasound images) a bigger influence of the smoothness term might improve the registration result. The use of a minimum spanning tree approach \cite{6471238} in the loss calculation to tackle sheering motion or the classification of an image into motion classes could be possible options.

\section{Conclusion}

In this paper we presented a DL based algorithm to calculate DVF for periodic motion tracking with 3D and 4D medical image datasets without the need of any training phase. For 3D datasets the algorithm is able to calculate forward and inverse transformation simultaneously. The algorithm was tested thoroughly with multiple public datasets. Compared to existing state of the art algorithms, our approach showed a competitive performance. The main benefits of our algorithm are that it is generic and can be used for different modalities and body regions and that no training data are needed in advance. The promising results suggest a possible utilisation in two different setting. First, as generic stand-alone algorithm for conventional image registration and motion tracking in e.g. radiotherapy treatment planning. Second, for studies focused on a specific organ and image modality the algorithm could be used without training phase until enough data are available and then be trained and thus specialised on a specific registration task to further increase registration accuracy.

\section{Acknowledgements}
We are thanking Richard Castillo, the Sunnybrook Health Sciences Centre as well as the Léon Bérard Cancer Center \& CREATIS lab for sharing the datasets that we used in this work to evaluate the presented algorithm.
We gratefully acknowledge the support of NVIDIA Corporation with the donation of the Titan Xp GPU used for this research. Further, we appreciate the valuable and constructive input from the editors and reviewers of the journal.

\textbf{Disclosure of Conflicts of Interest:} The authors have no relevant conflicts of interest to disclose.

\bibliographystyle{IEEEtran}
\bibliography{paper}

\begin{thebibliography}{10}
\providecommand{\url}[1]{#1}
\csname url@samestyle\endcsname
\providecommand{\newblock}{\relax}
\providecommand{\bibinfo}[2]{#2}
\providecommand{\BIBentrySTDinterwordspacing}{\spaceskip=0pt\relax}
\providecommand{\BIBentryALTinterwordstretchfactor}{4}
\providecommand{\BIBentryALTinterwordspacing}{\spaceskip=\fontdimen2\font plus
\BIBentryALTinterwordstretchfactor\fontdimen3\font minus
  \fontdimen4\font\relax}
\providecommand{\BIBforeignlanguage}[2]{{%
\expandafter\ifx\csname l@#1\endcsname\relax
\typeout{** WARNING: IEEEtran.bst: No hyphenation pattern has been}%
\typeout{** loaded for the language `#1'. Using the pattern for}%
\typeout{** the default language instead.}%
\else
\language=\csname l@#1\endcsname
\fi
#2}}
\providecommand{\BIBdecl}{\relax}
\BIBdecl

\bibitem{sotiras2013}
A.~Sotiras, C.~Davatzikos, and N.~Paragios, ``Deformable medical image
  registration: A survey,'' \emph{{IEEE} Transactions on Medical Imaging},
  vol.~32, no.~7, pp. 1153--1190, jul 2013.

\bibitem{Keszei2016}
A.~P. Keszei, B.~Berkels, and T.~M. Deserno, ``Survey of non-rigid registration
  tools in medicine,'' \emph{Journal of Digital Imaging}, vol.~30, no.~1, pp.
  102--116, oct 2016.

\bibitem{Oh2017}
S.~Oh and S.~Kim, ``Deformable image registration in radiation therapy,''
  \emph{Radiation Oncology Journal}, vol.~35, no.~2, pp. 101--111, jun 2017.

\bibitem{Tavares2014}
J.~M. R.~S. Tavares, ``Analysis of biomedical images based on automated methods
  of image registration,'' in \emph{Advances in Visual Computing}, G.~Bebis,
  R.~Boyle, B.~Parvin, D.~Koracin, R.~McMahan, J.~Jerald, H.~Zhang, S.~M.
  Drucker, C.~Kambhamettu, M.~El~Choubassi, Z.~Deng, and M.~Carlson, Eds.\hskip
  1em plus 0.5em minus 0.4em\relax Cham: Springer International Publishing,
  2014, pp. 21--30.

\bibitem{Tavares2014_2}
F.~P. Oliveira and J.~M.~R. Tavares, ``Medical image registration: a review,''
  \emph{Computer Methods in Biomechanics and Biomedical Engineering}, vol.~17,
  no.~2, pp. 73--93, 2014.

\bibitem{Tavares2015}
R.~S. Alves and J.~M. R.~S. Tavares, ``Computer image registration techniques
  applied to nuclear medicine images,'' in \emph{Computational and Experimental
  Biomedical Sciences: Methods and Applications}, J.~M. R.~S. Tavares and
  R.~Natal~Jorge, Eds.\hskip 1em plus 0.5em minus 0.4em\relax Cham: Springer
  International Publishing, 2015, pp. 173--191.

\bibitem{NIPS2015_5854}
M.~Jaderberg, K.~Simonyan, A.~Zisserman, and K.~Kavukcuoglu, ``Spatial
  transformer networks,'' in \emph{Advances in Neural Information Processing
  Systems 28}, C.~Cortes, N.~D. Lawrence, D.~D. Lee, M.~Sugiyama, and
  R.~Garnett, Eds.\hskip 1em plus 0.5em minus 0.4em\relax Curran Associates,
  Inc., 2015, pp. 2017--2025.

\bibitem{hu2018weakly}
Y.~Hu \emph{et~al.}, ``Weakly-supervised convolutional neural networks for
  multimodal image registration,'' \emph{Medical image analysis}, vol.~49, pp.
  1--13, 2018.

\bibitem{YANG2017378}
X.~Yang, R.~Kwitt, M.~Styner, and M.~Niethammer, ``Quicksilver: Fast predictive
  image registration – a deep learning approach,'' \emph{NeuroImage}, vol.
  158, pp. 378 -- 396, 2017.

\bibitem{Fan2018}
J.~Fan, X.~Cao, P.-T. Yap, and D.~Shen, ``Birnet: Brain image registration
  using dual-supervised fully convolutional networks,'' \emph{Medical Image
  Analysis}, vol.~54, pp. 193 -- 206, 2019.

\bibitem{Eppenhof2018}
K.~A.~J. Eppenhof and J.~P.~W. Pluim, ``Pulmonary {CT} registration through
  supervised learning with convolutional neural networks,'' \emph{{IEEE}
  Transactions on Medical Imaging}, pp. 1--1, 2018.

\bibitem{eppenhofSPIE}
K.~A.~J. Eppenhof, M.~W. Lafarge, P.~Moeskops, M.~Veta, and J.~P.~W. Pluim,
  ``{Deformable image registration using convolutional neural networks},'' in
  \emph{Medical Imaging 2018: Image Processing}, E.~D. Angelini and B.~A.
  Landman, Eds., vol. 10574, International Society for Optics and
  Photonics.\hskip 1em plus 0.5em minus 0.4em\relax SPIE, 2018, pp. 192 -- 197.

\bibitem{Vos2017}
B.~D. de~Vos, F.~F. Berendsen, M.~A. Viergever, M.~Staring, and I.~I{\v{s}}gum,
  ``End-to-end unsupervised deformable image registration with a convolutional
  neural network,'' in \emph{Deep Learning in Medical Image Analysis and
  Multimodal Learning for Clinical Decision Support}.\hskip 1em plus 0.5em
  minus 0.4em\relax Springer, 2017, pp. 204--212.

\bibitem{Vos2019}
B.~D. de~Vos, F.~F. Berendsen, M.~A. Viergever, H.~Sokooti, M.~Staring, and
  I.~I{\v{s}}gum, ``A deep learning framework for unsupervised affine and
  deformable image registration,'' \emph{Medical image analysis}, vol.~52, pp.
  128--143, 2019.

\bibitem{Dalca2018}
A.~V. Dalca, G.~Balakrishnan, J.~Guttag, and M.~R. Sabuncu, ``Unsupervised
  learning for fast probabilistic diffeomorphic registration,'' in
  \emph{Medical Image Computing and Computer Assisted Intervention -- MICCAI
  2018}, A.~F. Frangi, J.~A. Schnabel, C.~Davatzikos, C.~Alberola-L{\'o}pez,
  and G.~Fichtinger, Eds.\hskip 1em plus 0.5em minus 0.4em\relax Cham: Springer
  International Publishing, 2018, pp. 729--738.

\bibitem{Krebs2019}
J.~Krebs, H.~e~Delingette, B.~Mailhe, N.~Ayache, and T.~Mansi, ``Learning a
  probabilistic model for diffeomorphic registration,'' \emph{{IEEE}
  Transactions on Medical Imaging}, pp. 1--1, 2019.

\bibitem{Hongming2018}
H.~Li and Y.~Fan, ``Non-rigid image registration using self-supervised fully
  convolutional networks without training data,'' vol. 2018, 04 2018, pp.
  1075--1078.

\bibitem{Ferrante2018a}
E.~Ferrante, O.~Oktay, B.~Glocker, and D.~H. Milone, ``On the adaptability of
  unsupervised cnn-based deformable image registration to unseen image
  domains,'' \emph{Machine Learning in Medical Imaging}, jan 2018.

\bibitem{Maciejczyk2014}
A.~Maciejczyk, I.~Skrzypczyńska, and M.~Janiszewska, ``Lung cancer.
  radiotherapy in lung cancer: Actual methods and future trends,''
  \emph{Reports of practical oncology and radiotherapy : journal of Greatpoland
  Cancer Center in Poznan and Polish Society of Radiation Oncology}, vol.~19,
  pp. 353--360, Nov. 2014.

\bibitem{Oehlke2016}
O.~Oehlke \emph{et~al.}, ``Amino-acid pet versus mri guided re-irradiation in
  patients with recurrent glioblastoma multiforme (gliaa) - protocol of a
  randomized phase ii trial (noa 10/aro 2013-1).'' \emph{BMC cancer}, vol.~16,
  p. 769, Oct. 2016.

\bibitem{Sundar2009}
H.~Sundar, H.~Litt, and D.~Shen, ``Estimating myocardial motion by 4d image
  warping.'' \emph{Pattern recognition}, vol.~42, pp. 2514--2526, Nov. 2009.

\bibitem{vandemeulebroucke2011spatiotemporal}
J.~Vandemeulebroucke, S.~Rit, J.~Kybic, P.~Clarysse, and D.~Sarrut,
  ``Spatiotemporal motion estimation for respiratory-correlated imaging of the
  lungs,'' \emph{Medical physics}, vol.~38, no.~1, pp. 166--178, 2011.

\bibitem{Metz2011}
C.~T. Metz, S.~Klein, M.~Schaap, T.~van Walsum, and W.~J. Niessen, ``Nonrigid
  registration of dynamic medical imaging data using nd+ t b-splines and a
  groupwise optimization approach,'' \emph{Medical image analysis}, vol.~15,
  no.~2, pp. 238--249, 2011.

\bibitem{Peyrat2008}
J.-M. Peyrat, H.~Delingette, M.~Sermesant, X.~Pennec, C.~Xu, and N.~Ayache,
  ``Registration of 4d time-series of cardiac images with multichannel
  diffeomorphic demons,'' in \emph{Medical Image Computing and
  Computer-Assisted Intervention -- MICCAI 2008}, D.~Metaxas, L.~Axel,
  G.~Fichtinger, and G.~Sz{\'e}kely, Eds.\hskip 1em plus 0.5em minus
  0.4em\relax Berlin, Heidelberg: Springer Berlin Heidelberg, 2008, pp.
  972--979.

\bibitem{Peyrat2010}
J.-M. Peyrat, H.~Delingette, M.~Sermesant, C.~Xu, and N.~Ayache, ``Registration
  of 4d cardiac ct sequences under trajectory constraints with multichannel
  diffeomorphic demons.'' \emph{IEEE transactions on medical imaging}, vol.~29,
  pp. 1351--1368, Jul. 2010.

\bibitem{Wu2013}
G.~Wu, Q.~Wang, J.~Lian, and D.~Shen, ``Estimating the 4d respiratory lung
  motion by spatiotemporal registration and super-resolution image
  reconstruction,'' \emph{Medical Physics}, vol.~40, no.~3, p. 031710, feb
  2013.

\bibitem{Castillo2009}
R.~Castillo \emph{et~al.}, ``A framework for evaluation of deformable image
  registration spatial accuracy using large landmark point sets,''
  \emph{Physics in Medicine \& Biology}, vol.~54, no.~7, p. 1849, 2009.

\bibitem{Castillo2009a}
E.~Castillo, R.~Castillo, J.~Martinez, M.~Shenoy, and T.~Guerrero,
  ``Four-dimensional deformable image registration using trajectory modeling,''
  \emph{Physics in Medicine \& Biology}, vol.~55, no.~1, p. 305, 2009.

\bibitem{Radau2009}
P.~Radau, Y.~Lu, K.~Connelly, G.~Paul, A.~Dick, and G.~Wright, ``Evaluation
  framework for algorithms segmenting short axis cardiac mri,'' \emph{The MIDAS
  Journal-Cardiac MR Left Ventricle Segmentation Challenge}, vol.~49, 2009.

\bibitem{MiccaiLVChallenge}
``Cardiac mr left ventricle segmentation challenge,''
  \url{https://smial.sri.utoronto.ca/LV_Challenge/Home.html}, accessed:
  2019-04-25.

\bibitem{elastix2010}
S.~{Klein}, M.~{Staring}, K.~{Murphy}, M.~A. {Viergever}, and J.~P.~W. {Pluim},
  ``elastix: A toolbox for intensity-based medical image registration,''
  \emph{IEEE Transactions on Medical Imaging}, vol.~29, no.~1, pp. 196--205,
  Jan 2010.

\bibitem{plastimatch}
``plastimatch,'' \url{http://www.plastimatch.org/}, accessed: 2019-09-16.

\bibitem{FECHTER2017S757}
T.~Fechter, J.~Dolz, U.~Nestle, and D.~Baltas, ``Ep-1417: Clinical evaluation
  of a fully automatic body delineation algorithm for radiotherapy,''
  \emph{Radiotherapy and Oncology}, vol. 123, pp. S757 -- S758, 2017, {ESTRO}
  36, May 5-9, 2017, Vienna, Austria.

\bibitem{Ronneberger2015}
O.~Ronneberger, P.~Fischer, and T.~Brox, ``U-net: Convolutional networks for
  biomedical image segmentation,'' in \emph{Medical Image Computing and
  Computer-Assisted Intervention -- MICCAI 2015}, N.~Navab, J.~Hornegger, W.~M.
  Wells, and A.~F. Frangi, Eds.\hskip 1em plus 0.5em minus 0.4em\relax Cham:
  Springer International Publishing, 2015, pp. 234--241.

\bibitem{Dumoulin2016}
V.~Dumoulin and F.~Visin, ``A guide to convolution arithmetic for deep
  learning,'' \emph{arXiv e-prints}, Mar 2016.

\bibitem{Schnabel2001}
J.~A. Schnabel \emph{et~al.}, ``A generic framework for non-rigid registration
  based on non-uniform multi-level free-form deformations,'' in \emph{Medical
  Image Computing and Computer-Assisted Intervention -- MICCAI 2001}, W.~J.
  Niessen and M.~A. Viergever, Eds.\hskip 1em plus 0.5em minus 0.4em\relax
  Berlin, Heidelberg: Springer Berlin Heidelberg, 2001, pp. 573--581.

\bibitem{Roche2000}
A.~Roche, G.~Malandain, and N.~Ayache, ``Unifying maximum likelihood approaches
  in medical image registration,'' \emph{International Journal of Imaging
  Systems and Technology}, vol.~11, no.~1, pp. 71--80, 2000.

\bibitem{Vishnevskiy2014}
V.~Vishnevskiy, T.~Gass, G.~Sz{\'e}kely, and O.~Goksel, ``Total variation
  regularization of displacements in parametric image registration,'' in
  \emph{Abdominal Imaging. Computational and Clinical Applications},
  H.~Yoshida, J.~J. N{\"a}ppi, and S.~Saini, Eds.\hskip 1em plus 0.5em minus
  0.4em\relax Cham: Springer International Publishing, 2014, pp. 211--220.

\bibitem{sorensen1948method}
T.~S{\o}rensen, ``A method of establishing groups of equal amplitude in plant
  sociology based on similarity of species and its application to analyses of
  the vegetation on danish commons,'' \emph{Biol. Skr.}, vol.~5, pp. 1--34,
  1948.

\bibitem{Vishnevskiy2017}
V.~Vishnevskiy, T.~Gass, G.~Szekely, C.~Tanner, and O.~Goksel, ``Isotropic
  total variation regularization of displacements in parametric image
  registration,'' \emph{{IEEE} Transactions on Medical Imaging}, vol.~36,
  no.~2, pp. 385--395, feb 2017.

\bibitem{Parisi2019}
G.~I. Parisi, R.~Kemker, J.~L. Part, C.~Kanan, and S.~Wermter, ``Continual
  lifelong learning with neural networks: A review.'' \emph{Neural networks :
  the official journal of the International Neural Network Society}, vol. 113,
  pp. 54--71, May 2019.

\bibitem{6471238}
M.~P. {Heinrich}, M.~{Jenkinson}, M.~{Brady}, and J.~A. {Schnabel}, ``Mrf-based
  deformable registration and ventilation estimation of lung ct,'' \emph{IEEE
  Transactions on Medical Imaging}, vol.~32, no.~7, pp. 1239--1248, July 2013.

\end{thebibliography}

\end{document}